%% file: main_arxiv.tex
\documentclass{article}

\usepackage[margin=1in]{geometry}
\usepackage{natbib}

\usepackage[utf8]{inputenc}
\usepackage[T1]{fontenc}
\usepackage{hyperref}
\usepackage{url}
\usepackage{booktabs}
\usepackage{amsfonts}
\usepackage{amsmath}
\usepackage{nicefrac}
\usepackage{microtype}
\usepackage{xcolor}
\usepackage{graphicx}
\usepackage{multirow}

\newcommand{\jlens}{J-lens}
\newcommand{\ouro}{Ouro-2.6B}
\newcommand{\huginn}{Huginn-0125}
\newcommand{\qwen}{Qwen3.6-27B}

\title{Looped Transformers under the Jacobian Lens:\\Does the Global Workspace Survive Recurrence?}

\author{%
  Wenlong Wang\\
  Fin AI Research\\
  \texttt{wenlong.wang@intercom.io}
  \and
  Fergal Reid\\
  Fin AI Research\\
  \texttt{fergal.reid@intercom.io}
}

\date{}

\begin{document}

\maketitle

\begin{abstract}
Recent work identifies a mid-depth band of verbalisable, causally potent
representations in a standard feedforward transformer --- a functional analogue
of a global workspace. Whether the same workspace functionality emerges when
depth is implemented through recurrence rather than a stack of distinct layers
remains unknown. Looped and
depth-recurrent transformers provide a direct test of this question because
they reuse the same weights across depth. We extend the Jacobian lens to
iterated architectures using a virtual-unrolling adapter. We apply the full
workspace suite --- lens fitting, readout, and eleven causal experiment
families --- to \ouro{} (48 layers looped 4 times, deeply supervised) and
\huginn{} (a 4-layer core recurred 16 times, trained for latent reasoning),
using \qwen{} (64 untied layers) as the standard baseline. We find that a
workspace forms in the iterated part of each architecture, but that recurrence
changes how it can be accessed. Ouro reconstructs workspace content in every
loop, and linear
transport cannot carry that content across loop boundaries; writes and
ablations must therefore span every remaining loop. Huginn carries content
forward across all sixteen recurrences, while reads, writes, and ablations act
only within a sliding window of roughly two recurrences. Whether newly
injected content can be verbalised tracks explicit per-iteration supervision;
whether existing content can be steered does not.
\end{abstract}

\input{sections/01_intro}
\input{sections/02_background}
\input{sections/03_method}
\input{sections/04_transport}
\input{sections/05_loops}
\input{sections/06_geometry}
\input{sections/07_causal}
\input{sections/08_discussion}

\bibliographystyle{plainnat}
\bibliography{refs}

\appendix
\input{sections/appendix}

\end{document}

%% file: sections/01_intro.tex
\section{Introduction}
\label{sec:intro}

The global-workspace account of cognition holds that flexible behaviour depends
on a limited-capacity workspace whose contents are broadcast widely, are
reportable, and are open to directed modulation
\citep{baars1988,dehaene2011,mashour2020}.
\citet{workspace2026} operationalise this picture for language models: using a
\emph{Jacobian lens} --- per-layer linear maps fitted to the model's own mean
input--output Jacobians --- they identify, in a standard transformer, a
mid-depth band of residual-stream representations that are verbalisable,
causally necessary for behaviour, and selectively involved in flexible
rather than automatic computation.

Their discussion, however, flags a structural disanalogy with the biological
theory. In the brain, workspace dynamics are thought to be realised by
\emph{recurrent} processing --- ignition, sustained reverberation, broadcast
over re-entrant loops. A standard transformer has no built-in recurrence
within a forward pass: information transits a fixed stack of untied layers
exactly once. The authors observe that a feedforward network can emulate a
recurrent one for as many steps as it has layers --- a recurrent model being
the special case in which weights are tied across layers --- but state
plainly that they do not know whether depth in a transformer plays a role
similar enough to recurrence in a brain to enable the same workspace
functionality \citep{workspace2026}.

Looped and depth-recurrent transformers allow us to test this question
directly because they re-apply tied weights across iterations. Trained,
competitive examples of both architectures are now available, so we apply the
same \jlens{} and experimental suite used in the original study. We examine
two architecturally distinct members of the weight-tied family, with a
standard transformer as the baseline:

\begin{itemize}
  \item \textbf{\ouro{}} \citep{ouro2025}: 48 unique layers applied 4 times
    (virtual depth 192), trained with \emph{deep supervision} --- every loop
    end is directly optimised to be decodable;
  \item \textbf{\huginn{}} \citep{geiping2025}: a 4-layer core recurred a
    test-time-variable number of times between a 2-layer prelude and a 2-layer
    coda, trained for latent reasoning with truncated backpropagation and
    \emph{no} explicit deep supervision (we fit at $r{=}16$, virtual depth 68);
  \item \textbf{\qwen{}}: 64 untied layers --- the same model family as the
    original study, reproduced under our harness so all comparisons share one
    protocol.
\end{itemize}

We extend the Jacobian lens to iterated architectures with a
\emph{virtual-unrolling} adapter that presents the weight-tied computation
graph as a deep feedforward stack, so the original fitting and readout
machinery applies unchanged (\S\ref{sec:method}). We then re-run the workspace
suite: lens readout across virtual depth, the structural analysis of the
transport maps themselves, the Neuronpedia demonstration metrics, and eleven
causal experiment families spanning verbal report, steering, flexible
generalisation, ablation, and introspection
(\S\ref{sec:transport}--\S\ref{sec:causal}).

\paragraph{Looped transformers do have a global workspace --- but recurrence
reshapes its interfaces.}
Both weight-tied models exhibit the workspace signature of the original
study: their content is verbalisable, causally steerable, and selectively
engaged. Depth-as-recurrence therefore does support workspace function. The
two architectures differ, however, in how that workspace can be accessed.
Linear transport fails at Ouro's checkpointed loop boundaries and beyond a
$\sim$2-recurrence horizon on Huginn. Writes and ablations follow the same
structure: they must span every remaining loop on Ouro but remain confined to
the transport horizon on Huginn. The two training regimes also place
maintenance and verbalisation on opposite sides of the boundary. This paper
makes the following contributions:

\begin{itemize}
  \item \textbf{A method for lensing iterated models.} A virtual-unrolling
    adapter that presents the weight-tied computation graph to the original
    fitting machinery as a deep feedforward stack, giving exact
    per-virtual-layer lenses, together with a validation suite for the
    silent failure modes recurrence introduces (\S\ref{sec:method}).
  \item \textbf{Transport structure.} Mean-Jacobian transport is bounded by
    iteration structure: Ouro's supervised checkpoints decode and re-encode
    content, destroying the workspace component even where the bulk state
    survives, and Huginn's transport horizon is $\sim$2 recurrences for
    every fitted target. Weight tying leaves an exact symmetry in the fitted
    artefacts --- independently fitted Huginn lenses at matched offsets
    agree at cosine 0.9991 --- so one iteration's lens transfers to all
    (\S\ref{sec:transport}).
  \item \textbf{Two maintenance regimes.} Both models refine iteratively
    with per-task depth, but Ouro \emph{re-derives} its workspace content
    each loop through the checkpoint, while Huginn \emph{carries it
    forward} --- a persistence that injection-path interventions show to be
    continuous recomputation from the re-injected input, not a memory
    buffer (\S\ref{sec:loops}, \S\ref{sec:prelude}). An occupancy
    analysis --- measured with the original study's sparse concept
    inventory, which we implement --- controls for lens distance and yields
    the same geometry: four workspace bands on Ouro, and on Huginn a flat
    plateau across the recurrent stack with a single step into the coda
    (\S\ref{sec:geometry}).
  \item \textbf{Architecture-specific causal interfaces.} In Ouro, an
    intervention must be maintained across all remaining loops because each
    loop tends to reconstruct the original content. In Huginn, writes and
    ablations act only within the transport horizon, where the last lens
    window alone is exactly as effective as all of them. The two verbal-report
    experiments produce different results: Huginn reroutes self-computed
    content into words nearly perfectly (98\%) yet never verbalises externally
    injected content (0/97), which only explicitly deep-supervised Ouro
    achieves (\S\ref{sec:causal}).
  \item \textbf{Validation of lens readouts.} A readout taken beyond a lens's
    validated horizon does not read the state. If lens targets are unevenly
    spaced, apparent trends across depth reflect the instrument rather than
    changes in the model's representations. And readout claims can be
    estimator-specific: two results obtained with the demonstration's
    simpler top-8 readout do not survive re-measurement with the concept
    inventory (Appendix~\ref{app:pursuit}), and a causal null obtained
    with a single-direction ablation disappears at ten directions. We
    document these failure modes and the controls that guard against them
    (\S\ref{sec:window}, \S\ref{sec:occupancy}, \S\ref{sec:ablation}).
\end{itemize}

%% file: sections/02_background.tex
\section{Background and related work}
\label{sec:background}

\subsection{The Jacobian lens}
\label{sec:jlens-recap}

Following \citet{workspace2026}, the Jacobian lens at source layer $v$ with
target layer $t$ is the mean input--output Jacobian of the intervening
computation,
\begin{equation}
  J_v \;=\; \mathbb{E}\!\left[\frac{\partial h_t}{\partial h_v}\right],
  \label{eq:jlens}
\end{equation}
estimated over a corpus of forward passes. The lens readout at layer $v$
unembeds the linearly transported state, $\mathrm{unembed}(h_v J_v^{\top})$,
and the \emph{lens direction} for a token $u$ is the normalised pullback
$d = J_v^{\top} u / \lVert J_v^{\top} u\rVert$ of its unembedding vector.
Directions support causal edits: injection ($h \leftarrow h + \alpha d$),
ablation ($h \leftarrow h - (h \cdot d)\,d$), and clamp-transfer swaps between
two tokens' directions. The original work shows that on a standard transformer
the set of layers where these readouts are interpretable and these edits are
potent forms a contiguous mid-depth band with global-workspace-like functional
properties. The Jacobian lens is related to earlier affine readout methods,
including the logit lens \citep{nostalgebraist2020} and tuned lens
\citep{belrose2023}. Unlike those methods, it is derived from the model's own
local linearisation rather than fitted to a decoding objective. Its
\emph{failure} to transport can therefore carry architectural information
(\S\ref{sec:transport}).

\subsection{Looped and depth-recurrent transformers}
\label{sec:looped-taxonomy}

We use \emph{looped transformer} broadly for any transformer whose forward
pass applies a weight-tied block more than once --- the lineage of the
Universal Transformer \citep{dehghani2019} and of parameter-sharing encoders
\citep{lan2020}, recently revived for scaling latent reasoning
\citep{ouro2025,geiping2025} and studied theoretically as a model of iterated
computation \citep{giannou2023}. Three design axes matter for what follows:

\begin{description}
  \item[Loop scope.] Ouro loops the \emph{entire} stack: all 48 layers form
    one weight-tied block, and each iteration re-applies the whole
    transformer. Huginn loops only a 4-layer \emph{core}, sandwiched between
    an untied 2-layer prelude and 2-layer coda that run exactly once --- so
    part of Huginn's depth is a fixed feedforward frame around the
    recurrence, and its recurrent unit is far smaller than its virtual
    depth suggests.
  \item[Iteration count.] Fixed at training time (Ouro: 4 loops, always) or
    variable at test time (Huginn: sampled during training, chosen at
    inference; we use $r{=}16$).
  \item[Supervision of intermediate states.] Ouro attaches the LM head to
    every loop end and trains all of them (\emph{deep supervision}); Huginn
    trains only the post-coda output, with truncated backpropagation through
    the recurrence (mean unroll 8).
\end{description}

Table~\ref{tab:models} summarises the three models. Huginn at $r{=}16$ is a
close depth-and-compute match to \qwen{} (68 vs.\ 64 virtual layers), which
makes the untied/tied comparison unusually direct.

\begin{table}[t]
  \centering
  \caption{The three models. ``Virtual depth'' is the number of layer
    applications in one forward pass; sharing is what distinguishes the
    families.}
  \label{tab:models}
  \small
  \begin{tabular}{lllll}
    \toprule
    & Params & Physical layers & Virtual depth & Intermediate supervision \\
    \midrule
    \qwen{}   & 27B  & 64 (untied)              & 64  & none \\
    \ouro{}   & 2.6B & 48, looped $\times 4$    & 192 & deep supervision at every loop end \\
    \huginn{} & 3.5B & $2 + 4 + 2$, core $\times r$ & 68 ($r{=}16$) & none (truncated BPTT, mean 8) \\
    \bottomrule
  \end{tabular}
\end{table}

\subsection{Related work}
\label{sec:related}

\paragraph{Global workspace theory and AI systems.} Global workspace theory
\citep{baars1988,dehaene2011,mashour2020} motivates the functional suite we
inherit; \citet{butlin2023} propose indicator properties derived from such
theories for assessing AI systems, with recurrent processing as a recurring
ingredient. We make no claim about consciousness: we use the workspace suite
only as a functional characterisation and ask how its outcomes depend on
architectural recurrence.

\paragraph{Readout probes.} Logit lens \citep{nostalgebraist2020}, tuned lens
\citep{belrose2023}, and the Jacobian lens \citep{workspace2026} differ in how
their affine readout maps are obtained: fixed, trained, or derived from the
model, respectively. Probing of
weight-tied models specifically is nascent: \citet{lu2025latentcot} decode
Huginn's recurrent states with the logit lens and a coda lens, finding that
the two disagree systematically across core blocks and little evidence of
an interpretable latent chain of thought, and \citet{blayney2026mechanistic}
characterise looped models' iteration dynamics through hidden-state
geometry, attention patterns, and linear probes. Neither fits transport
maps between virtual layers, which is what the workspace suite requires.

\paragraph{Iterated computation.} Looped transformers have been analysed as
programmable computers \citep{giannou2023} and as learners of iterative
algorithms \citep{yang2024looped}; Ouro and Huginn demonstrate the recipe at
language-model scale \citep{ouro2025,geiping2025}. Interpretability work on
these models has so far decoded states through fixed unembedding-based
probes --- the logit lens in the Huginn release itself \citep{geiping2025},
logit and coda lenses across its recurrent stack \citep{lu2025latentcot} ---
or analysed iteration dynamics mechanistically
\citep{blayney2026mechanistic}; to our knowledge no prior work fits
transport-based lenses across loop boundaries or runs a causal workspace
suite on a looped model.

%% file: sections/03_method.tex
\section{Method: virtual unrolling}
\label{sec:method}

\subsection{The lens on an unrolled graph}

The lens definition (Eq.~\ref{eq:jlens}) extends to weight-tied models by
generalising ``layer'' to \emph{virtual layer}: the $k$-th firing of a
physical block is its own node in the unrolled computation graph. For Ouro,
$v = 48\cdot\text{step} + \text{layer}$; for Huginn, prelude and coda blocks
fire once and the $i$-th core block's $k$-th firing is $v = 2 + 4k + i$. Our
adapter presents the unrolled graph to the original fitting machinery by
replacing the model's layer list with virtual-block proxies whose forward
hooks dispatch only on their assigned firing --- without this, a standard
activation recorder silently keeps only the \emph{last} iteration's state.
The remaining gradients are computed automatically through the between-loop
normalisation and the tied weights, so $J_v$ is exact for the unrolled graph,
not an approximation to it. Fitting through a recurrence also exposes failure
modes that do not exist on feedforward stacks and that corrupt results without
raising any error; we document three such silent-failure traps, and the
validation suite that guards against them, in Appendix~\ref{app:traps}
(library-compatibility patches for loading the two looped models in
Appendix~\ref{app:compat}).

\subsection{Fit protocol}

All lenses share one protocol --- the original study's recipe --- so
cross-model comparisons never mix estimators: 1000 WikiText-103 prompts of 128
tokens (identical prompt set for all three models), first 16 positions
skipped, parameters frozen, per-source mean Jacobians accumulated by the
unmodified fitting code. Table~\ref{tab:fit} summarises. Convergence was
monitored as the relative shift of the running mean (final shift $\le 0.016$
on all Huginn lenses; $\approx 5\times 10^{-3}$ on the Qwen band-entry lens).
All fits and experiments ran on one to three NVIDIA B200 GPUs in bf16.

\begin{table}[t]
  \centering
  \caption{Lens-fitting summary. Every lens uses the same 1000-prompt corpus
    and estimator; only the (source, target) geometry differs. The twelve
    Huginn gap-fill lenses put a target at every remaining recurrence end,
    completing the depth tiling used for readout in \S\ref{sec:occupancy};
    the causal suite (\S\ref{sec:causal}) uses the four $r{=}4/8/12/16$
    lenses only.}
  \label{tab:fit}
  \footnotesize
  \setlength{\tabcolsep}{4pt}
  \begin{tabular}{llll}
    \toprule
    & Sources & Targets & Wall-clock \\
    \midrule
    \qwen{}   & all 63 layers; L0--23 (band entry) & final; L24 & $\sim$3\,h; $\sim$2\,h \\
    \ouro{}   & all 191 virtual layers & virt.\ final + loop ends (v47/95/143) & $\sim$1\,h + 1.6\,h \\
    \huginn{} & all 67 virtual layers  & coda end + $r{=}4/8/12$; gap-fill $r{=}1$--$15$ & $\sim$7.3\,h + per-target \\
    \bottomrule
  \end{tabular}
\end{table}

\subsection{Why a lens \emph{family}, not a lens}
\label{sec:lens-family}

Our initial fitting scheme treated the unrolled model as one long,
weight-tied transformer and used its final virtual state as the lens target,
as on a standard transformer. This scheme fails as a readout instrument on both iterated
models. Ouro's final-target lens reads punctuation throughout loops 1--3 ---
not because those loops are empty, but because the transport to the final
target dies crossing loop boundaries (\S\ref{sec:transport}); Huginn's
transport horizon is roughly two recurrences regardless of target.

We therefore fit a family of lenses whose targets are the outputs of
individual loops or recurrences: every loop end on Ouro
(v47/95/143/191) and at the recurrence ends $r{=}4/8/12/16$ on Huginn
(Table~\ref{tab:fit}). Note that we change only the lens target, not the
fitting procedure: an input--output Jacobian is well defined for any upstream
source and downstream target in the unrolled computation graph. The causal
experiments in \S\ref{sec:causal} validate the resulting lens family
empirically: interventions along its token directions change what the models
verbalise and how they answer in the predicted, concept-specific ways,
providing the same form of causal validation used by
\citet{workspace2026}. On Ouro the choice of target is principled, not
merely convenient: loop ends are exactly the states deep supervision trains
the LM head to read, so $\mathrm{unembed}(h J^{\top})$ is a proper readout at
every target, and the per-loop lenses reveal that the semantic content the
final-target lens missed was present from loop 2 all along. On Huginn the
targets tile the depth so that every source lies within some lens's
transport horizon: those four lenses carry the causal suite
(\S\ref{sec:causal}), and for readout we complete the family to a target at
\emph{every} recurrence end, putting every source within one recurrence of
its own target (\S\ref{sec:occupancy}). Throughout the paper, every readout is
therefore taken through the lens \emph{whose target the source can actually
reach}, and every cross-loop or cross-model comparison is
\emph{distance-controlled}: readouts are compared at matched offsets from
their own targets, where transport attenuation is matched by construction.
Matching source-to-target distance prevents differences in transport
attenuation from being mistaken for architectural effects.

%% file: sections/04_transport.tex
\section{Transport structure: what the lens artefacts reveal}
\label{sec:transport}

We first study the fitted transport maps themselves --- how far they carry
information, what symmetries they inherit from weight tying, and how they
compose. Throughout, we distinguish two things a transport can carry. The
\emph{bulk} of the state is the residual vector as a whole --- the
thousands-dimensional aggregate that matrix-level statistics (transport norm,
KL, top-1 agreement) measure. The \emph{semantic workspace component} is the
low-dimensional slice of that vector that encodes nameable concepts --- the
part the workspace phenomena actually live in, measured by whether a specific
concept remains recoverable after transport. The distinction matters because
the two behave completely differently: bulk transport collapses with distance
in the same way across all three architectures, while the workspace component
and compositional structure are where recurrence makes a difference.

\subsection{Bulk collapse with distance is universal}
\label{sec:bulk}

We use bulk statistics to control for source-to-target distance. Transport
weakens with distance in all three architectures, so cross-architecture
comparisons must be made at matched distances. A mean Jacobian is only
informative where individual prompts' Jacobians agree. Near its target they
do (per-prompt
$\lVert J\rVert/\sqrt{d} \approx 1.3$--$1.5$ on Ouro, mean $1.24$ one block
out); across tens of layers the response direction becomes prompt-dependent
and the average cancels ($\lVert J\rVert/\sqrt{d} = 0.13$--$0.3$ beyond one
Ouro loop). This decline is similar across architectures: at distance
39 --- the standard model's own working band-to-target span --- \qwen{}'s
aggregate transport is just as dead as Ouro's at matched distance (top-1
agreement with the final output: 2\%, $\lVert J\rVert/\sqrt{d}=0.46$,
inside Ouro's 0.14--0.53 range across its target series). This collapse does
not mean that the lens is unusable: the same distance-39 transport that fails
every aggregate test still carries a specific latent concept to its target
(Table~\ref{tab:boundary}, first row). At long distances, useful readout
depends not on the bulk response but on a small set of semantic directions
that remain consistent across prompts. Unembedding a direction-washed,
near-zero transport of $h$ yields an unspecific vector \emph{regardless of the
content $h$ carries}; the few directions on which prompts' Jacobians agree
survive the averaging that cancels everything else.

\paragraph{Methodological consequence.} Aggregate faithfulness metrics
($\lVert J\rVert$, KL, top-1 agreement against the final output) are not
usability tests for a lens at range, and we never use them as such. We regard
a readout as valid when it moves a probe token's rank against a $\sim V/2$
chance baseline and its direction supports a causal intervention; bulk
numbers appear only in matched-distance comparisons.

\subsection{The semantic component does not cross Ouro's loop boundary}
\label{sec:boundary}

Once bulk attenuation is controlled, the architectural effect can be isolated
clearly.
We track a prompt-absent latent concept --- \emph{Italy}, from the riddle
``Fact: The currency used in the country shaped like a boot is\ldots'' ---
through transports whose endpoints are both independently proven to contain
it (Table~\ref{tab:boundary}).

\begin{table}[t]
  \centering
  \caption{Semantic transport at matched bulk quality, tracked with the
    riddle prompt ``Fact: The currency used in the country shaped like a boot
    is\ldots''.
    The latent concept \emph{Italy} never appears in the prompt, and both
    endpoints of every row are independently verified to represent it.
    ``Carries \emph{Italy}?'' asks whether the concept survives the transport
    itself: after mapping the source state through the fitted transport, is
    \emph{Italy} still recoverable at the target (top-ranked in the
    transported readout, with a concept-vs-control projection above chance)?
    The cross-boundary transport (last row) has \emph{better} bulk statistics
    than the standard model's working case (first row), yet the concept does
    not arrive.}
  \label{tab:boundary}
  \small
  \begin{tabular}{lccc}
    \toprule
    Transport & Distance & State cosine & Carries \emph{Italy}? \\
    \midrule
    \qwen{} band $\to$ target (within stack) & 39 & 0.19 & yes \\
    \ouro{} band $\to$ own loop end (within loop) & 11--23 & 0.3--0.5 & yes \\
    \ouro{} loop-2 band $\to$ loop-3 band (across boundary) & 32--60 & 0.31--0.45 & \textbf{no} \\
    \bottomrule
  \end{tabular}
\end{table}

Across the boundary the \emph{Italy} projection drops to $\sim$0.1 of the
target state's value with no concept-vs-control discrimination, while bulk
magnitude survives at 0.27--0.46: the workspace component is destroyed roughly
$3\times$ faster than the state as a whole. The mechanism is what happens at
Ouro's loop ends. Deep supervision trains every loop-end state to be directly
decodable by the LM head --- we call these states \emph{checkpoints},
in the sense of supervised way-points \emph{within} a single forward pass
(not saved model weights). Because the training
loss reads each checkpoint directly (logit-lens KL at loop ends: 1.7/0.5/0.1
nats for loops 1--3, top-1 agreement up to 92\%; one block \emph{short} of a
checkpoint, at virtual layer 190, decodability falls to 0.8 nats and 71\% ---
states are readable exactly where training made them readable), each loop
\emph{decodes}
its conclusions into a token-aligned checkpoint state, and the next loop
\emph{re-encodes} them nonlinearly into its own band coordinates. Successive
loops therefore exchange information through token-aligned checkpoint states
rather than preserving the same workspace vectors across the boundary.

Is content destroyed because the model iterates at all, or because of this
decode/re-encode step? Huginn separates the two possibilities. It also
iterates a weight-tied block, but its recurrence boundary has no supervised
decode step: the raw latent state flows unchanged into the next iteration. If
iteration itself destroyed content, Huginn's should degrade the same way.
Instead, workspace content computed by recurrence 2--3 remains readable
through all 16 recurrences (\S\ref{sec:loops}). The destruction is therefore
a property of the checkpoint interface, not of looping.

\subsection{Weight tying induces symmetries in the lens artefacts}
\label{sec:symmetry}

Weight tying admits two interpretations of these models: a looped transformer
is ``a deeper transformer that happens to reuse weights'', or it is ``one
programme run repeatedly''. The two interpretations make different predictions about the fitted
transports. If only depth position matters, per-iteration structure should be
invisible in them; if the model re-runs one programme, the transports should
repeat when viewed from each iteration's own frame. Settling this also has a
practical implication: if the transports repeat, a lens fitted at one iteration
transfers to the others, and the loop-relative, distance-controlled
comparisons this paper relies on (\S\ref{sec:lens-family}) are legitimate
rather than convenient. The results support each interpretation in a
different coordinate frame.

\paragraph{Absolute frame: no trace of tying.} Cosine similarity between flattened
$J_v$ pairs on Ouro decays smoothly with virtual distance (0.989 adjacent,
0.804 at $\Delta{=}48$, 0.087 at $\Delta{=}190$) with no block structure at
multiples of 48; same-physical-layer pairs are no more similar than the
all-pairs baseline (0.676 vs.\ 0.711). Towards the final output, the looped
model is indistinguishable from a generic 192-layer network: what a layer
application does to the eventual output is set by where it sits in the
unrolled computation, not by which weights it uses.

\begin{figure}[t]
  \centering
  \includegraphics[width=\textwidth]{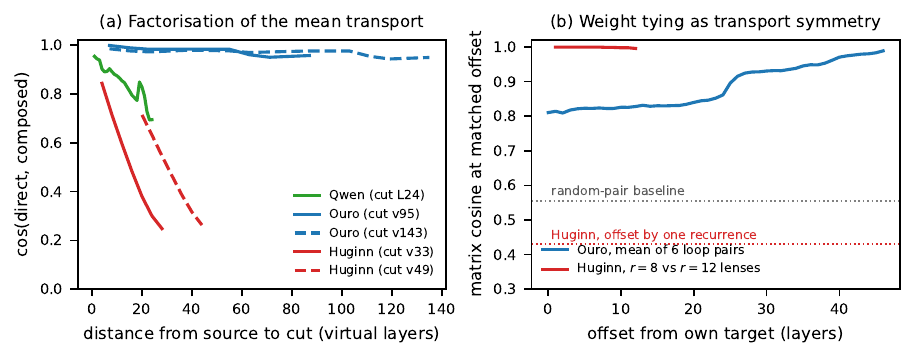}
  \caption{The lens artefacts carry the architecture's signature.
    \textbf{(a)} Factorisation of the mean transport through a mid-depth cut,
    $\cos(J_{v\to t}, J_{w\to t}J_{v\to w})$ vs.\ source--cut distance.
    Intuitively, a high cosine means the long journey decomposes into two
    independent legs --- the state at the cut summarises everything the rest
    of the computation needs. Deep-supervised Ouro decomposes almost
    perfectly at any range (its loops are Markovian: each hands the next a
    self-contained summary); the standard transformer decomposes only
    approximately, degrading with span; Huginn barely decomposes at all ---
    its recurrences stay entangled across any cut
    (\S\ref{sec:factorisation}).
    \textbf{(b)} Similarity of independently fitted transports compared at
    matched offsets from their own targets. A weight-tied core does the same
    thing at every iteration, and the fitted lenses inherit that repetition:
    exactly on Huginn (cosine 0.9991 --- the $r{=}8$ and $r{=}12$ lenses are
    interchangeable), approximately on Ouro (0.81--0.99, rising towards the
    target). Dotted lines show what dissimilar transports look like: random
    pairs, and Huginn's own lenses misaligned by one recurrence
    (\S\ref{sec:symmetry}).}
  \label{fig:transport}
\end{figure}

\paragraph{Loop-relative frame: strong to exact symmetry.} Compared at
matched offsets from \emph{their own} targets --- which matches attenuation by
construction --- the per-iteration transports repeat. On Ouro, matched-offset
matrices across loops have mean cosine 0.885 (0.932 inside the workspace
band), against 0.716 offset-mismatched and 0.555 for random pairs; their
top-32 singular subspaces overlap at 0.86/0.83 (random baseline 0.016), and
pairwise similarity rises with loop convergence (loops 3--4: 0.988). On
Huginn the symmetry is essentially \emph{exact}: independently fitted lenses
at matched offsets agree with matrix cosine \textbf{0.9991} ($r{=}8$ vs.\
$r{=}12$ targets; 0.43 when offset by one recurrence;
Figure~\ref{fig:transport}b). The weight-tied core
implements literally the same mean transport at every recurrence.

These results confirm both interpretations, each in its own frame. Towards
the final output, tying is invisible
(same-physical-layer similarity 0.676 vs.\ all-pairs 0.711) --- as a
contributor to the eventual answer, a looped transformer really is ``a deeper
transformer that happens to reuse weights'', and what matters is when a layer
runs, not which weights it runs. Within each iteration's own frame, the
transports repeat far above every baseline (0.885--0.9991 vs.\ 0.43--0.716)
--- the model really does ``run one programme repeatedly''. The practical
consequence follows from the second reading: a lens fitted at one iteration
transfers to the others (on Huginn, exactly), which legitimises the
loop-relative comparisons used throughout and means full-depth lens coverage
of a recurrent model costs one iteration's fit, not $r$ of them.

\subsection{Factorisation orders the architectures}
\label{sec:factorisation}

Per prompt, the chain rule guarantees $J_{v\to t} = J_{w\to t}\,J_{v\to w}$;
for the \emph{means} the factorisation
$\mathbb{E}[AB] \approx \mathbb{E}[A]\,\mathbb{E}[B]$ is not guaranteed, and
whether it holds measures how much prompt-dependent correlation crosses the
cut. The three architectures order cleanly
(Table~\ref{tab:factorisation}, Figure~\ref{fig:transport}a).

\begin{table}[t]
  \centering
  \caption{Mean-transport factorisation through a mid-depth cut:
    $\cos\!\big(J_{v\to t},\, J_{w\to t} J_{v\to w}\big)$.}
  \label{tab:factorisation}
  \small
  \begin{tabular}{llc}
    \toprule
    & Cut & cos(direct, composed) \\
    \midrule
    \ouro{}   & supervised loop ends (v95, v143) & 0.94--1.00 \\
    \qwen{}   & band entry, L24                  & 0.69--0.95, rising towards the cut \\
    \huginn{} & recurrence ends (v33, v49)       & 0.25--0.84, rising towards the cut \\
    \bottomrule
  \end{tabular}
\end{table}

Ouro's near-perfect factorisation is evidence of deep supervision: the
checkpoint truncates cross-cut correlations, making the loop-to-loop process
approximately Markovian --- which is the same mechanism that destroys
persisting content (\S\ref{sec:boundary}). Huginn shows the complementary
pattern: its continuous latent flow preserves
prompt-dependent correlations across boundaries (so content persists), and
the product of means, which keeps only the uncorrelated part, under-predicts
the true transport. The standard transformer sits between: approximate
factorisation with a correlation term that grows with span.

\paragraph{Practical corollary.} On checkpointed models, per-loop lenses can
be fitted independently (memory scales with one loop's span) and composed to
recover any long-range transport --- the composition matches the end-to-end
fit. On continuous-recurrence models it cannot; end-to-end fitting is
required.

\subsection{The input-injection path}
\label{sec:prelude}

Huginn re-injects the prelude's output at every recurrence, and the lens sees
it: transport norm from the prelude is distance-\emph{independent} towards
every target (0.26--0.37 from prelude layer 0, 0.54--0.77 from layer 1) while
mid-recurrence sources collapse below 0.06 within 4--5 recurrences. The
recurrent state is better read as a working memory continuously re-grounded
in a fixed input encoding than as a residual stream --- likely also why its
state-to-state mean transport decorrelates so fast. Ouro shows a weak
analogue (its loop-1-target lens rises again where its source is the
embedding layer). This path matters causally: it is the interface that makes
mid-stream interventions decay unless refreshed (\S\ref{sec:causal}).

These results do not establish whether looped transformers should re-inject
the input encoding. Huginn and Ouro differ simultaneously in
injection, supervision, and training recipe, so nothing here attributes
Huginn's content persistence to the injection path specifically --- \S%
\ref{sec:boundary} attributes the \emph{destruction} on Ouro to its
checkpoint interface, which is a different mechanism. In the one model that
has the path, it is the only long-range route the mean transport preserves,
and it coexists with --- but is not shown to cause --- stable content
maintenance. Two inference-time experiments,
both hooks on the adapter's
$\mathrm{linear}(\mathrm{cat}[\text{state}, \text{embeds}])$ input, settle
the maintenance question without retraining. The two hypotheses predict
opposite outcomes: if the recurrent state were a memory buffer, content
computed by recurrence 7 would already sit in the state --- which these
hooks leave untouched --- and a late cut to the injection should not
matter. \emph{Attenuation}: scaling the
injected embedding by $\alpha$ from recurrence $k$ onwards destroys the
output at $\alpha \le 0.25$ for \emph{every} onset --- 0/8 prompts recover
the baseline next token even when only the last four of sixteen recurrences
are attenuated --- while $\alpha = 0.75$ is nearly harmless (7--8/8). The
lens shows why: at $\alpha = 0$, $k = 8$, the latent answer's rank holds at
its baseline level through recurrence 7 (median 11) and collapses within one
recurrence of the cut (2{,}749 at recurrence 8, then $\sim$40{,}000 ---
chance). Maintenance depends on re-grounding: content
does not survive even one recurrence beyond its last refresh.
\emph{Injection swap}: replacing the injected embedding with a
matched-length prompt's (three pairs differing in one word, both directions)
from recurrence $k$ onwards flips the final output to the \emph{new} input's
own answer in all 30 configurations, including $k = 12$; the old answer's
rank decays within $\sim$2 recurrences of the swap and the new one reaches
near rank 1 within $\sim$4. The state is therefore best described as
continuous recomputation from the injected encoding with a
$\sim$2-recurrence memory, not a memory buffer --- the same constant
that bounds the mean transport (\S\ref{sec:bulk}), now measured causally.
Natural content ``persists'' across all sixteen recurrences
(\S\ref{sec:window}) because the input that regenerates it is re-injected at
every one. A definitive design answer still needs controlled training runs
(matched small looped models with and without injection), which is beyond
this paper's scope.

%% file: sections/05_loops.tex
\section{What the iterations compute}
\label{sec:loops}

Section~\ref{sec:transport} characterised how information moves; this section
uses the lens family to ask what the iterations actually do with it. The main
result is that both models perform iterative refinement with per-task depth,
but they maintain their intermediate results in different ways. A
concept crossing an Ouro loop boundary is destroyed and re-derived by the
next loop, while on Huginn a concept computed by recurrence 2--3 stays
readable through all sixteen recurrences, sustained by the re-injected
input encoding rather than by the recurrent state (\S\ref{sec:prelude}).

\subsection{Ouro: iterative refinement, mostly done by loop 3}
\label{sec:refinement}

Because Ouro's checkpoints are trained to be decodable
(\S\ref{sec:boundary}), the plain logit lens is close to a calibrated
instrument \emph{at loop ends} (0.1--1.7 nats from the final distribution,
60--92\% top-1 agreement), and cross-loop timing claims can rest on it.
Reading the loop ends over WikiText: loop-3's output already matches the
final output for 92\% of tokens; loop 4 changes the top-1 for only
$\sim$8\%. The model's own early-exit gate --- a trained component that uses
no lens at all --- corroborates the same timeline, assigning exit probability
$[0.03, 0.14, 0.34, 0.49]$ across the four loops.

The per-loop lens family shows which representations change. Tracking latent
concepts through each loop's own band (a distance-controlled comparison,
\S\ref{sec:lens-family}):

\begin{itemize}
  \item \textbf{Easy fact} (``The capital of the country where the Eiffel
    Tower stands is the city of\ldots''): \emph{Paris} is top-1 at the end of
    loop 1 and every later loop end --- the remaining loops maintain, not
    compute.
  \item \textbf{Riddle} (``Fact: The currency used in the country shaped
    like a boot is\ldots''): loop-1's band reads only the literal surface
    (\emph{boots, footwear, shoe}); \textbf{loop-2's band already reads
    \emph{Italy}}; loops 3--4 maintain it. The readout identifies a correct
    token that is absent from the prompt, indicating that the transport
    retains semantic information.
  \item \textbf{Arithmetic} ($47 + 38$): the answer settles in the mid-band
    readout only in loop 4, though the checkpoint state shows
    \emph{eighty} from loop 2, suggesting that a partial answer forms early
    but does not settle fully until later.
\end{itemize}

Task difficulty maps onto the number of loops before the band contains the
answer. Three measurements with different failure modes support this account
of convergence:

\begin{enumerate}
  \item \textbf{Per-loop \jlens{} readouts.} Measurement: the rank of a
    prompt-absent probe token (e.g.\ \emph{Italy}) in
    $\mathrm{unembed}(h_v J_{v \to \text{loop end}}^{\top})$, at each loop's
    band layers. Depends on our fitted transport.
  \item \textbf{Logit lens at checkpoints.} Measurement: KL and top-1
    agreement between $\mathrm{unembed}(h_{\text{loop end}})$ and the model's
    final output distribution, over WikiText. This is the LM head applied
    directly to the loop-end state --- exactly the operation the
    deep-supervision loss trained --- and involves no transport.
  \item \textbf{The early-exit gate.} Measurement: the model's own trained
    stopping signal --- the learned gate that decides at each loop end whether
    to halt --- logged per token, with no readout of any kind on our part. Its
    mean exit distribution over WikiText is $[0.03, 0.14, 0.34, 0.49]$ across
    loops 1--4: the model itself would stop at loop 3 for a third of tokens.
\end{enumerate}

The first depends on the lens estimator, the second only on direct
unembedding, and the third on neither --- it is an internal component of the
model, trained end-to-end, that we merely observe. All three agree that
convergence is mostly complete by loop 3 with per-task depth, so the picture
cannot be an artefact of any single estimator.

\subsection{Huginn: a sliding window, and content that persists}
\label{sec:window}

Huginn has no absolute band. Each lens in the family reads reliably only in
the one-to-two recurrences before its own target: for the prompt ``Fact: The
capital city of the country famous for sushi is'', the latent answer
\emph{Tokyo} is rank 1 at recurrence steps 2--3 through the $r{=}4$
lens, step 7 through $r{=}8$, steps 10--11 through $r{=}12$, and steps 14--15
through the primary lens. The readable region \emph{slides with the target}
--- transport is distance-limited (\S\ref{sec:bulk}), and workspace geometry
follows the lens, not the layer index.

Because every window reads the \emph{same} content, the latent answer ---
computed by recurrence 2--3 --- persists in the state through all 16
recurrences. This is the direct opposite
of Ouro's maintenance strategy. Ouro's checkpoints destroy the workspace
component at each boundary and the next loop re-derives it
(\S\ref{sec:boundary}); Huginn's raw-latent handoff carries it forward,
regenerated from the re-injected input at every recurrence
(\S\ref{sec:prelude}). The two architectures show the same relationship
between task difficulty and the iteration at which the answer becomes
readable: the
answer to ``\ldots the country famous for sushi\ldots'' (\emph{Tokyo}) is
readable from recurrence 2, ``The currency used in the country home to the
Eiffel Tower\ldots'' (\emph{French}) by recurrence 3, and the boot riddle of
\S\ref{sec:refinement} arrives late and equivocally --- it is genuinely hard
for a 3.5B model. Thus, in both architectures, harder tasks settle later in
the iterative computation, while ``destroy-and-re-derive vs.\ carry-forward''
separates the two training regimes.

\begin{figure}[t]
  \centering
  \includegraphics[width=\textwidth]{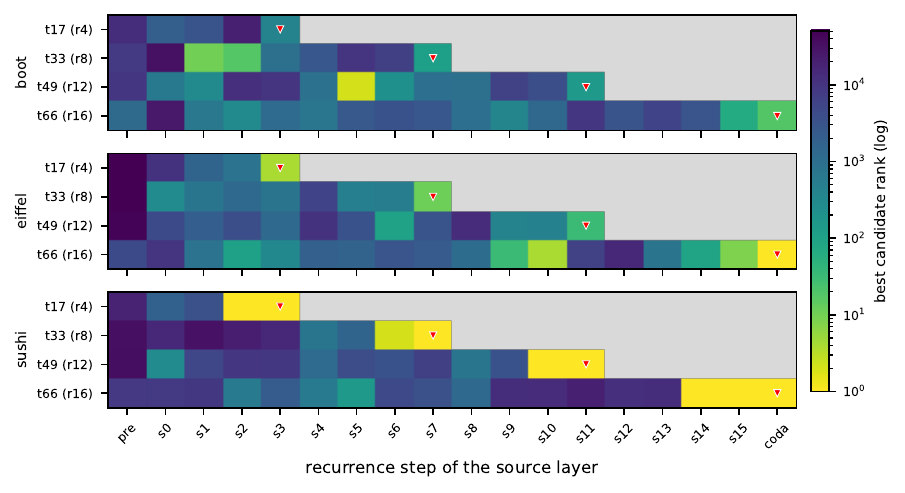}
  \caption{Band-map heat strip on Huginn: best rank among the expected latent
    candidates (log colour scale; brighter is better) at every source layer,
    bucketed by recurrence step, for the four original lenses (rows; red
    marker = each lens's own target) on three carrier prompts (panels). All
    rows read the \emph{same} $r{=}16$ forward pass; grey cells lie outside a
    lens's fitted source range. The readable region slides with the lens
    target (\S\ref{sec:window}): each lens resolves content only in the
    one-to-two recurrences before its target, and the model answers the
    riddles with graded difficulty (\emph{sushi} readable from recurrence 2
    everywhere, \emph{boot} only weakly and late).}
  \label{fig:bandmap}
\end{figure}

Figure~\ref{fig:bandmap} shows both results at once as a heat strip over the
same run: each lens's readable region (bright cells) hugs the one-to-two
recurrences before its own target, and the four regions tile the depth axis
while reading the same latent content.

\subsection{Uniform, weak decodability at every depth --- without deep supervision}
\label{sec:implicit}

Directly unembedding Huginn's recurrent state (no transport at all) is
informative at every depth --- uniformly, and weakly. Over the 8-prompt
suite, the logit lens's top-1 token equals the model's actually emitted
next token at 15--25\% of positions, flatly across the whole recurrent stack
(virtual layers 14--65; final coda state: 63\%). The right comparison is the
other models mid-computation: Qwen never exceeds 9\% anywhere in L8--40, and
Ouro sits at 1--7\% mid-loop, spiking to 64--71\% only at its supervised
checkpoints. Huginn is thus the only model whose states decode above the
mid-computation floor at \emph{every} depth, although with only a quarter of
the confidence at Ouro's checkpoints. The plausible mechanism is \emph{implicit}
deep supervision:
training samples the recurrence count from a broad distribution (mean 32), so
across the corpus the single LM head is applied to states of every depth,
exerting the same decodability pressure that Ouro's explicit per-step loss
applies. This result shows that logit-lens decodability is \emph{not} evidence
of explicit deep supervision: head pressure at depth in any form produces it.
It also predicts that Huginn should verbalise workspace content well despite
lacking Ouro's supervision, which \S\ref{sec:causal} confirms.

%% file: sections/06_geometry.tex
\section{Workspace geometry across architectures}
\label{sec:geometry}

The readouts of \S\ref{sec:loops} rest on target-token \emph{ranks} for
hand-picked probes. This section re-measures the workspace geometry with an
independent family of metrics, computed from recorded forwards of the
original study's demonstration protocol --- 8 chat prompts, 128-token
greedy completions, and both lenses read at every (layer, position)
(Appendix~\ref{app:demo}). Occupancy is measured with the study's own
concept-inventory estimator --- a gradient-pursuit
\citep{blumensath2008gradient,nanda2024progress} reconstruction of the
state, described in the original but with no released implementation; ours
is specified in Appendix~\ref{app:pursuit}. We use these metrics to test
whether the same structure appears under measurement procedures with
different failure modes.

\subsection{Occupancy recovers the band, loop, and window structure}
\label{sec:occupancy}

\begin{figure}[!tp]
  \centering
  \includegraphics[width=\textwidth]{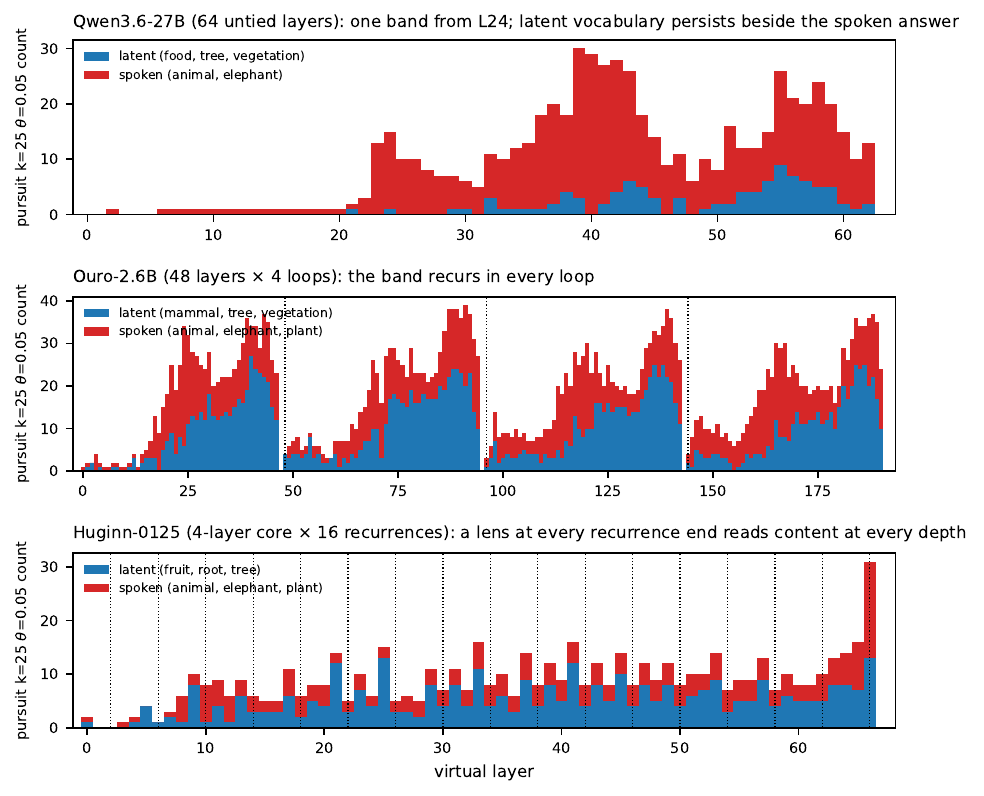}
  \caption{Count-by-layer occupancy of the \jlens{} concept inventory
    (pursuit at $k = 25$, coefficient threshold $0.05\,\lVert h \rVert$) on
    the animal prompt, the same for all three models (verbatim prompt and
    replies in Appendix~\ref{app:transcripts}; the top-8 counterpart is
    Figure~\ref{fig:cbl-top8}). Bars are \emph{stacked},
    not overlaid: blue is the latent class (read but never written;
    singular/plural variants merged), red the spoken class (in the
    transcript) drawn on top, so total bar height is the two classes'
    combined count. Dotted lines mark loop boundaries (Ouro, every 48
    physical layers; the rise into L41--46 before each boundary is the
    decode into that loop's supervised checkpoint, \S\ref{sec:transport})
    and recurrence boundaries (Huginn, every 4 virtual layers between the
    2-layer prelude and coda). The periodic patterns have different sources:
    Ouro's four bands are a \emph{model} property (content forms
    and dies within each loop, with real boundaries between loops), while
    Huginn's sawtooth is an \emph{instrument} property --- each lens
    window's count rises towards its own lens target at the mean-J
    transport horizon, tiling content that persists continuously across
    all sixteen recurrences (\S\ref{sec:window}).}
  \label{fig:cbl}
\end{figure}

At each (layer, position) cell, pursuit reconstructs the state as a sparse,
non-negative combination of unit-normalised \jlens{} atoms $a_\tau =
J_v^{\top} u_\tau$. A token \emph{occupies} the cell when its atom is among
the first 25 selected with a coefficient above $0.05\,\lVert h \rVert$; its
occupancy at a layer is the number of such cells. This measure requires no
ground-truth answer. Appendix~\ref{app:pursuit} specifies the algorithm,
reports robustness across the full $(k, \theta)$ grid, and cross-checks the
results with an independent estimator: the demonstration's top-8 softmax
readout. The two estimators select substantially different tokens.
Figure~\ref{fig:cbl} plots occupancy across virtual depth for the animal prompt
(``I'm thinking of the largest land animal. What do they eat?''). This is the
only demonstration prompt that all three models answer directly and correctly
(verbatim prompt and replies appear in Appendix~\ref{app:transcripts}). We
divide the tracked tokens into \emph{latent} tokens, which occur in the
readout but not the transcript, and \emph{spoken} tokens, which occur in the
transcript; singular and plural variants are merged. The latent class contains
related concepts such as \emph{food} (the category asked about), \emph{tree}
(unsaid by all three models), \emph{mammal}, \emph{vegetation}, \emph{fruit},
and \emph{root}.

This metric recovers the band, loop, and window structure described in
\S\S\ref{sec:transport}--\ref{sec:loops}. On
Qwen, the latent vocabulary is absent below L20 (0 slots; an absence
carried by the coefficient sizes rather than by which atoms are
selected, Appendix~\ref{app:pursuit}), fills the L24--48 band beside the
spoken answer (42 latent vs 326 spoken), and persists into the top of
the stack (37 vs 112 at L55+). On Ouro, in-band (physical L24--40)
occupancy exceeds out-of-band occupancy as a \emph{per-layer rate} in all
four loops and for both token classes (loop 4: 22.3 vs.\ 16.6 slots per
layer; the band contains 17 of each loop's 47 lensed layers). Per-layer
rates are necessary because the out-of-band region contains nearly twice as
many layers, causing the comparison to fail on raw totals in the late loops.
The relationship also weakens at $k = 50$, where the still-rising variance
curve (Appendix~\ref{app:pursuit}) indicates that the additional selected
atoms are no longer matched by additional J-space content. On Huginn, read through the completed
tiling described below (every source within one recurrence of its
covering lens target, so lens distance is held constant across depth),
total occupancy rises through the prelude and first few recurrences and then
plateaus over the middle of the stack: the eleven consecutive
one-recurrence windows spanning v17--v60 vary only between 31 and 44
slots, against 5 in the first window. It then jumps in the coda window
(92 slots, $2.4\times$ the plateau mean). After controlling for lens
distance, the same concepts remain readable at a similar rate across the
recurrent stack rather than accumulating with depth. The only marked increase
occurs in the coda state consumed by the LM head.

The other demonstration prompts provide more specific examples. On the
currency riddle (``what currency was used in the boot-shaped country before
2002?''), Ouro correctly answers ``the Italian lira''. The unspoken
alternative \emph{euro}, a tempting surface association between 2002 and
currency, peaks in loop~2 and then declines from 259 to 188. Meanwhile,
\emph{Italy}, the intermediate concept used to derive the correct answer,
increases monotonically from 177 to 390. Across all twelve grid settings, the
representation therefore shifts
from the rejected alternative towards the intermediate concept used by the
correct answer. The same analysis identifies why Huginn answers the riddle
incorrectly (``was the Euro''): the country never enters J-space
(\emph{Italy / Italian / lira / lire}: 5 slots across all 67 virtual
layers, and never more than 9 anywhere on the grid), so the
salient surface association (2002 + currency $\to$ euro) dominates. This is
the occupancy counterpart of the two-hop weakness measured behaviourally by
the probe swap (5/90, \S\ref{sec:causal}). Qwen's inventory also includes
concepts across languages: its occupied atoms include the Chinese token
for \emph{Italy} (102 slots), alongside the English \emph{Italy} (344) and
\emph{currency} (225). English-string token classes cannot capture this
concept mass. On the sport prompt, Qwen's latent sport concepts occupy 123
band slots but none in the final eight layers, while spoken tokens retain 116
slots. The late layers therefore retain the emitted words while dropping the
unspoken topic alternatives. As with the absence below L20, the zero is due
to coefficient sizes (Appendix~\ref{app:pursuit}).

Completing Huginn's lens family to a full
tiling --- a target at \emph{every} recurrence end --- makes occupancy
nonzero at every virtual layer, directly confirming the persistence
result of \S\ref{sec:window}: the same concepts are readable wherever a
lens's transport reaches. However, the original partial tiling makes content
appear to accumulate through the recurrences. The rising curve is an artefact
of unequal source-to-target lens distance rather than a property of the model
(Appendix~\ref{app:tiling}).

\subsection{Decodability, quantified from the probability field}
\label{sec:calibration}

The dumps also record, for every (layer, position), the probability the
lens's decoded distribution assigns to the token the model actually emits
next --- $\mathrm{softmax}(\mathrm{unembed}(h))$ for the logit lens,
evaluated at the true next token and averaged over positions and prompts
(top-8-truncated, hence a lower bound). Chance on a $\sim$150k vocabulary is
$\sim$$10^{-5}$, which sets the scale. This field calibrates the
\emph{confidence} of the decodability established by rank in
\S\ref{sec:loops}: Ouro's logit lens reaches 0.61--0.65 at \emph{every} loop
end --- the checkpoint state, decoded directly, already gives the model's
actual next token two-thirds probability, i.e.\ nearly the final layer's
confidence, which is deep supervision made visible. Huginn sits on a flat
plateau of 0.02--0.045 across the recurrent stack from recurrence 3 onward
(virtual layers 14--65, with a stable within-recurrence profile that peaks
at each recurrence's last core layer), jumping to 0.47 only at the final
coda state v66 --- the state the LM head consumes. The plateau is three orders of
magnitude above chance and matches the rank picture of
\S\ref{sec:implicit} (top-1 hit rate 15--25\% uniformly across the same
depths), yet sits $20\times$ below Ouro's checkpoints --- implicit
supervision buys readability at every depth, not checkpoint-grade
confidence. Qwen's
mid-band is near zero through the logit lens (0.003 at L36) and an order of
magnitude better through the \jlens{} (0.019) --- on a standard transformer,
mid-stack states are not next-token-decodable without transport. The
three-model ordering --- Ouro checkpoints $\gg$ Huginn everywhere-weakly $>$
Qwen mid-band $\approx 0$ --- also appears in the demonstration's own
probability field, without a separate set of target-token probes.

%% file: sections/07_causal.tex
\section{The causal suite: eleven intervention families}
\label{sec:causal}

The preceding analyses are observational: they read states that the model
computed on its own. We next test the inferred directions using causal
interventions. The original study released eleven
behavioural experiment families (writes, ablations, and readout probes over
workspace directions); we reimplemented them in a single shared harness and
ran the full suite identically on all three models. Two protocol choices
matter. First, all interventions write or ablate along lens directions
$d = \mathrm{normalise}(J_v^{\top} u_t)$ --- the artefact under study is the
intervention operator. Second, each model uses its own validated interface:
Qwen intervenes over the band L24--48 through the final-target lens; Ouro
clamps the bands of loops 2--4 with each loop's own lens; Huginn clamps the
last three lens windows. Table~\ref{tab:suite} condenses the results
(full tables and token-position conventions in
Appendix~\ref{app:tables}).

\begin{table}[t]
  \centering
  \caption{The eleven-family causal suite. Each family is scored by the
    original study's own metric (second column gives its reported range where
    one exists). Bold marks the best model per family. The pattern the
    section unpacks: the three models succeed on \emph{different} families,
    and the split tracks training interface, not scale.}
  \label{tab:suite}
  \footnotesize
  \setlength{\tabcolsep}{4pt}
  \begin{tabular}{llccc}
    \toprule
    Family & Study & \qwen{} 27B & \ouro{} 2.6B & \huginn{} 3.5B \\
    \midrule
    Verbal report (swap $\to$ top-5) & 59--88\% & 69\% & 80\% & \textbf{98\%} \\
    Introspection (injected concept) & reportable & null & \textbf{top-1 31\%} & null (0/97) \\
    Modulation (focus/ctrl/suppress) & ordered & .38/.07/.07 & \textbf{.77/.39/.26} & .28/.15/.19 \\
    Top-down summoning & evokes concept & $\sim$0 & $\sim$0 & $\sim$0 \\
    Two-hop probe swap & 54--70\% & \textbf{52\%} (base 63\%) & 29\% (base 48\%) & 6\% (base 28\%) \\
    Flexible generalisation (swap) & 40\% & \textbf{42\%} & 38\% & 13\% (base 44\%) \\
    Selectivity: language & expl.\ $\gg$ auto & \textbf{+1.00} & +0.25 & +0.38 \\
    Selectivity: line count & iff relevant & 0.09@1 & \textbf{1.00@1} & 0.09@1 \\
    Ignition (10--90 width) & sharp & \textbf{0.10} & 0.55 & none in-window \\
    Capacity (words held) & $\sim$25 & 0.7 & \textbf{2.4} & 1.8 \\
    Dual task (concepts survive) & survive & 6/21 & \textbf{17/21} & 7--8/21 \\
    \bottomrule
  \end{tabular}
\end{table}

\subsection{Writes must span every remaining iteration}
\label{sec:writes}

The flexible-generalisation experiment swaps a concept mid-reasoning: on a
prompt whose answer routes through an intermediate concept, the operator
clamps the workspace state's projection onto the source concept's lens
direction over to the target concept's, and scores whether the model's answer
flips to the one implied by the injected concept. On Qwen the swap succeeds
--- the answer flips to the injected concept's implication --- in 42\% of
the 192 trials, matching the original study's 40\%. On Ouro, \emph{where}
the clamp is applied decides
everything: clamping the bands of all remaining loops (2--4) reaches 38\% ---
statistically indistinguishable from Qwen --- but any single loop alone
falls well short: 21\% for loop 2, 29\% for loop 3, 25\% for loop 4. The
loop-2-only result is the causal
counterpart of \S\ref{sec:refinement}: later loops do not passively copy the
patched state forward --- they \emph{re-derive} content from the raw context
and partially restore the original answer. A write into an iterated model
must therefore be maintained across every remaining loop to counteract this
reconstruction. Huginn exhibits a complementary constraint: clamping the last
window alone gives results identical to
clamping all three windows, while the second-to-last window alone gives 3\%
--- the $\sim$2-recurrence transport horizon of \S\ref{sec:window} bounds
\emph{writes} exactly as it bounds reads, because a lens direction is only a
valid coordinate system near its own target. Although the original content
persists across recurrences (\S\ref{sec:window}), an intervention expressed in
one lens's coordinates does not.

The loop-4-only drop (25\%) cannot be re-derivation --- no loop runs after
it. Per-trial logit-lens rank trajectories of both answers, logged at every
loop end, show the two extreme conditions fail for \emph{opposite} reasons.
Unclamped, the original answer is settled early: median rank 1 by the end
of loop 2 on every category (1--3 at loop 1). Under loop-2-only clamping,
failed trials show the patch \emph{taking} and then being undone --- the
injected answer reaches median rank 7 at the loop-2 checkpoint but decays to
44 by loop 4 while the original recovers to rank 1: overwritten by
re-derivation. Under loop-4-only clamping, failures look entirely different:
the original answer already sits at rank 1 at loop-4 \emph{entry} and the
clamp lifts the injected answer only to median rank 11 --- a displacement
failure against consolidated content, with no re-derivation involved. Loop
3 combines the two mechanisms: it has one fewer re-derivation pass than loop
2 and less consolidated opposition than loop 4. Its result falls between the
two extremes at 29\% (CI 23--36), exactly as the two-mechanism account
requires. On all-loops-clamped \emph{successes} the
original answer is driven from rank 1 to median rank 294 by loop 4 --- the
sustained clamp is precisely what prevents the recovery seen in the
single-loop conditions. The category split matches: countries (same-category direction cosine 0.36,
well-separated) succeed $\sim$30/48 in every condition, while months
(cosine 0.67) and animals (0.39) fail in opposite directions across the two
single-loop conditions (months 0 early vs.\ 12 late; animals 9 vs.\ 6, but
24 with all loops).

\subsection{Ablation follows the same interfaces as writes}
\label{sec:ablation}

We test two widths of ablation. The first removes a single concept's lens
direction ($h \leftarrow h - (h \cdot d)\,d$ at the intervention layers); the
second removes the ten-dimensional span of the inventory atoms that best
reconstruct the state (\S\ref{sec:occupancy}). For both, we measure whether an
answer that was correct before the intervention remains correct.

The single-direction results show that the effect is specific to the concept.
Ablating the prompt's argument concept degrades all three
models comparably (answer kept: Qwen 72\%, Ouro 71\%, Huginn 61\%), while
a control that repeats the operation with an unrelated word's direction
never changes the answer (100\% kept for all three models). The behavioural
effect therefore depends on which direction is removed rather than on
removing an arbitrary direction, confirming that the lens direction carries
the concept. The interface pattern observed for writes in
\S\ref{sec:writes} also appears under ablation: Ouro's
loop-2-only ablation is a \emph{complete} null (100\% kept --- the later
loops re-derive the ablated content through the checkpoint, mirroring the
21\%-vs-38\% write pattern), and on Huginn ablating the last window alone is
\emph{exactly} equivalent to ablating every window (17/28 kept in both
conditions). Removing the \emph{answer} token's direction reveals a further
ordering across models: it leaves Qwen at 53\% kept, Ouro at 19\%, and Huginn
at 0\% (the
answer's rank collapses to $\sim$$10^{3.5}$). This ordering follows the
proximity of each model's states to token space (\S\ref{sec:implicit},
\S\ref{sec:calibration}): removing the answer direction has the largest
effect on directly decodable representations, whereas Qwen's more distributed
mid-stack retains roughly half of its correct answers.

The ten-direction ablation tests which results depend on the interface and
which depend on the width of the ablation. At each intervention (layer,
position), we project out the span of the ten inventory atoms that best
reconstruct that state (bases orthonormalised and frozen from
an uninstrumented forward pass, protocol in Appendix~\ref{app:pursuit});
a matched control removing the span of ten fixed random-token atoms leaves
answers 87--97\% intact on every model. The effects therefore reflect removal
of the inventory subspace rather than the loss of any ten dimensions. For
Huginn, the last window alone is as destructive as all windows together
(0/28 vs.\ 1/28 kept), while the second-to-last window remains a null (27/28).
Thus, the window equivalence becomes even stronger under the wider ablation.
Ouro behaves differently: removing the ten-direction subspace at loop~2 alone
leaves 6/42 answers (14\%), whereas removing a single concept direction left
42/42. Re-derivation through the checkpoint restores a deleted direction but
not the state's wider J-space component. A loop-2 ablation therefore has no
behavioural effect only when it is narrow enough for the next loop to
reconstruct, and single-direction ablation understates what that band carries. Qwen changes in
the same direction, but less strongly: 64\% of answers remain correct under
the ten-direction ablation, compared with 72\% for the argument direction and
97\% for the control.

\subsection{Verbalising new content requires explicit deep supervision;
  steering existing content does not}
\label{sec:verbalisation}

The verbal-report and introspection experiments produce contrasting results.
Huginn is the \emph{best} of the three at verbal report --- swap a concept
the model has already computed and is about to emit, and it reports the
injected concept in its top-5 at 98\% (124/126) --- yet a \emph{null} at
introspection: injecting a concept the model was not computing never makes
it reportable (0 of 97 concepts reach rank $\le$ 100 at any injection
strength), even though the injections measurably move ranks (median 40{,}666
$\to$ 16{,}258 at the strongest). Ouro is the only model that wins the
second task --- injected concepts become top-1 reportable at 31\% --- and
also the only model whose training explicitly applies the LM head at every
iteration. This refines and resolves the prediction left open in
\S\ref{sec:implicit}: implicit head pressure (Huginn's sampled
recurrence depths) suffices to make \emph{existing} content steerable and
reportable --- rerouting what the model already computed, where Huginn's
exact loop-invariant transport (\S\ref{sec:symmetry}) and persistence
(\S\ref{sec:window}) make the final windows an unusually clean write surface
--- but making \emph{new}, injected content verbalisable tracks explicit
per-iteration supervision, and nothing else in the suite substitutes for
it. Scale is not a substitute: the 27B Qwen model is also null at
introspection (its best injected concept reaches only rank 173).

\subsection{What the nulls mean: readable is not usable, and inert content
  is invisible by construction}
\label{sec:nulls}

The suite's failures are as informative as its successes, but they are not
one kind of failure. We distinguish three kinds: \emph{capability
floors} (the model cannot do the task, so the intervention has nothing to
show), \emph{readout bounds} (the content is there, but the lens cannot
decode enough of it), and one \emph{instrument blind spot} (content the lens
cannot see even in principle). Each changes what a null is evidence of.

\paragraph{Capability floors.} The language-selectivity task shows the model
a passage and asks a question about it (naming its author); the selectivity
signature is that the passage's \emph{language} --- never asked about ---
should be readable in the band only when the task makes it relevant. Huginn
shows exactly that signature (language label hit@5 1.00 when task-relevant
vs.\ 0.62 when incidental) --- yet answers the author question itself 0/8.
The workspace demonstrably holds and gates the content, but the 3.5B model
lacks the capability to use it. Huginn's poor probe-swap and
flexible-generalisation rows carry the same caveat --- its no-intervention
task baselines (28\% and 44\%) are the lowest of the three models --- so
those nulls are partly ``cannot do the task'', not ``has no workspace''.

\paragraph{Readout bounds.} The capacity task asks the model to hold a list
of words in mind (four 20-word blocks, probed at comma positions) and counts
how many remain retrievable from the band. All three models score far below
the original study's $\sim$25 words --- but the ordering is \emph{inverse}
to parameter count (Ouro 2.4 $>$ Huginn 1.8 $>$ Qwen 0.7), which rules out
``small models store less''. It instead matches how \emph{decodable} each
model's band is (\S\ref{sec:calibration}): the measure is bounded by what
the readout can extract, not by what the state holds. The inventory
estimator of \S\ref{sec:occupancy} makes this bound directly visible:
scoring the same word lists by pursuit reconstruction instead of the rank
criterion roughly doubles every plateau --- Qwen 0.7 $\to$ 2.0, Ouro 2.4
$\to$ 4.9, Huginn 1.8 $\to$ 3.6 words --- while preserving the
inverse-to-scale ordering and remaining a factor of five below the
original's $\sim$25. The measured capacity moves with the estimator, as an
instrument bound must; and under a coefficient threshold of
$0.05\,\lVert h\rVert$ Qwen's plateau collapses to 0.1, consistent with
how little of its state the atoms explain (4.3\% variance,
\S\ref{sec:occupancy}). Our capacity figures
are facts about the instruments' reach, not the models' memory.

\paragraph{The instrument blind spot.} Top-down summoning names a category
label and asks the model to bring an absent concept to mind; the measure is
whether that concept becomes readable in the band above matched foils. It is
$\sim$0 on all three models, and in the strongest sense: the summoned
concept scores no better than foils, and asking vs.\ not asking changes
nothing --- the manipulation never takes hold. What summoning shares with
capacity is that both probe content that is \emph{not driving the current
next token} --- a concept merely held in mind, a word list waiting to be
recalled. The lens is structurally blind there. $J_v =
\mathbb{E}[\partial h_{\text{target}} / \partial h_v]$ is built entirely
from ``how does this state influence the target'' --- so any content that
is not currently influencing the next token contributes nothing to $J$, and
the lens maps it to (near) zero regardless of whether it is present. The
instrument therefore cannot distinguish ``not held'' from ``held but idle'';
the
limitation is the same for all three models, which is why these failures
are the only uniform ones in Table~\ref{tab:suite}. Any study using
Jacobian artefacts as a data modality inherits it: the lens sees the
workspace's \emph{working} content, not its inventory.

\subsection{Robustness: reasoning SFT does not reorganise the workspace}
\label{sec:thinking}

Everything above is measured on a pre-trained checkpoint; one might worry
the workspace interfaces are fragile to further training. Re-running the
complete pipeline --- lens fitting, the structural analyses, and
the full causal suite --- on Ouro-2.6B-Thinking, the reasoning SFT of our
base checkpoint ($\sim$8.3M supervised examples) shows that they are not. Every architecture-level
signature reproduces within a few points --- writes must still span every
remaining loop, the loop-2-only ablation is again a near-null, and
introspection is point-identical at top-1 31\% --- while task baselines dip
slightly and intervention effects move with them. The pursuit-based
measurements reproduce alongside the rest: the capacity plateau is 5.1
words (base 4.9) and the loop-2-only subspace ablation keeps 11\% (base
14\%). The two exceptions are
themselves informative: the language-selectivity gate collapses to always-on
($\Delta$ $+$0.25 $\to$ $+$0.00), and ignition sharpens (10--90 width 0.55
$\to$ 0.20). Appendix~\ref{app:thinking} gives the full comparison.

%% file: sections/08_discussion.tex
\section{Conclusion}
\label{sec:conclusion}

\citet{workspace2026} left open whether depth in a transformer plays a
role similar enough to recurrence in a brain to enable the same workspace
functionality. For the weight-tied special case they identify, our functional
answer is yes. Both looped models exhibit the full
workspace signature --- content that is verbalisable, causally steerable by
writes and ablations along lens directions, and selectively engaged by task
demands. Re-applying the same weights therefore does not prevent a workspace
from forming. What recurrence changes is the workspace's
\emph{interfaces}: every route by which content enters, survives, or leaves
the workspace follows the iteration structure rather than absolute depth.

We find that the two models develop different representations at their
iteration boundaries, and these differences shape how their workspaces can be
accessed. At Ouro's supervised checkpoints, each loop's conclusions are
decoded into a token-aligned state and then re-encoded by the next loop. This
makes the loop-to-loop process nearly Markovian, destroys the workspace
component in transit, and forces any write to be maintained across every
remaining loop's
re-derivation (38\% spanning all loops vs.\ 21\% for one). Huginn's raw
latent handoff preserves content across all sixteen recurrences, but reads
and writes work only within a $\sim$2-recurrence horizon of a lens's own
target --- the last window alone is exactly as effective as all of them, for
writes and ablations alike. Weight tying also reduces the cost of fitting the
instrument: the fitted transports repeat across iterations (exactly on
Huginn, at cosine 0.9991), so full-depth lens coverage of a recurrent model
costs one iteration's fit rather than $r$ of them.

The report experiments separate two abilities that the original study's
standard transformer did not distinguish. Steering \emph{existing} content
into words needs only head pressure at depth in some form --- Huginn, with
no explicit deep supervision, is the best of the three models at verbal
report (98\%). Verbalising \emph{new}, injected content is a different
matter: of everything we tested, only explicit per-iteration supervision
predicted it. Ouro, the one model trained that way, is the only one that
reports an injected concept (top-1 31\%); Huginn never does (0/97), and the
27B Qwen model shows that untied scale cannot substitute for this
supervision. Among the models
tested, Huginn is also the closest transformer analogue of content held
across processing cycles: content computed by
recurrence 2--3 remains present, unverbalised, through every later
recurrence. The analogy has a sharp limit --- the injection-path experiments
show this persistence is continuous recomputation from a re-injected input
encoding, not a memory buffer; content does not survive even one
recurrence beyond its last refresh. Whether a looped architecture can hold
workspace content \emph{without} such re-grounding is a question for
controlled training runs, not for released checkpoints.

The contribution has two main limitations. The first concerns the instrument:
mean-Jacobian lenses are linearised summaries with a single-token concept
vocabulary. They stop reading the state beyond their validated horizon,
report the tiling rather than the model when lens families are unequally
spaced, and are structurally blind to content that is not driving the
current next token; readout counts depend on the estimator, and
single-direction interventions can understate what a subspace carries. Any
use of Jacobian artefacts as a data modality inherits these limits. Second,
our cross-architecture comparison uses one pre-trained checkpoint per family,
so architecture is confounded with training recipe and scale. Following
checkpoints throughout pretraining would test whether our conclusions are
stable over training and reveal when J-space emerges and how its geometry and
causal interfaces develop. A systematic sweep over Huginn's test-time
recurrence count also remains future work. Within those limits, the result is
consistent across three measurement families with different failure modes and
remains unchanged after reasoning SFT.

Despite their different interfaces, our results suggest that neither looped
model simply stores completed workspace content. Ouro's later loops recover
content after early interventions, while Huginn's output depends on continued
re-injection of the input throughout recurrence. Looped architectures are
adopted for their reasoning strength. An open question is whether this
continued computation contributes to that strength through built-in
self-correction, or is overhead that a better iteration interface could
remove. Two released checkpoints that differ simultaneously in supervision,
injection, and training recipe cannot separate these possibilities. The
instruments developed here make it possible to test them in controlled
training runs that vary only the boundary interface.

%% file: sections/appendix.tex
\section{Verification of reported numbers}
\label{app:harness}

Every quantitative claim in this paper is re-derived from the released raw
artefacts --- per-trial result records, run logs, and the readout dumps ---
by an automated verification suite included with the code release
(232 checks, plus a companion suite of 132 checks covering the
gradient-pursuit artefacts of Appendix~\ref{app:pursuit}). Each suite runs
with one command and audits every table and inline statistic. Most values are recomputed from per-trial records,
with percentages and 95\% Wilson intervals recomputed from raw counts. For
four experiment families the artefacts store only condition-level aggregates
(directed modulation, dual task, line-count selectivity, and the \qwen{}
verbal report --- marked \textsc{agg} in the tables below); for these the
suite verifies the reported value against the stored aggregate but cannot
independently re-derive it. Re-running those families with per-trial
recording is planned alongside experiment (iii) of the open question in
\S\ref{sec:writes}.

\section{Full causal-suite tables}
\label{app:tables}

Tables~\ref{tab:suite-full}--\ref{tab:ablation-full} expand the condensed
Table~\ref{tab:suite} to exact counts. Every value is re-derived by the
verification suite of Appendix~\ref{app:harness}; rows whose artefacts store
only aggregates are marked \textsc{agg}.

\paragraph{Token-position conventions.}
The lens itself carries no position index --- $J_v$ is a mean Jacobian,
marginalised over the fitting corpus's prompts and positions
(\S\ref{sec:method}) --- so position enters only at readout and intervention
time. Our default, matching the original study's stated protocol
\citep{workspace2026}, is that a direction intervention applies at
\emph{every token position} as well as at every listed virtual layer: when
\S\ref{sec:causal} says Ouro clamps the bands of loops 2--4, the clamp acts
on the whole sequence at every band layer of every remaining loop, and the
behavioural outcome is read from the final position's next-token
distribution. Verbal-report swaps, flexible-generalisation swaps, two-hop
probe swaps, and ablation all follow this default. Two families are
span-restricted: the summoning clamp-transfer edits only the stimulus's
token span --- more conservative than the original all-positions protocol,
since the edit reaches the answer position only through attention --- and
introspection injects over the token span of the user instruction, matching
the original's user-turn injection, with the effect read at the final
position. Rank-probe readouts (summoning, modulation, selectivity, capacity,
dual task) take the minimum lens rank over band layers $\times$ a
task-defined span: the stimulus or carrier-sentence tokens, every position
after the probed passage, or the comma positions of the capacity list.

\begin{table}[t]
  \centering
  \caption{The full causal suite with exact counts. Columns follow
    Table~\ref{tab:suite}; ``base'' is the no-intervention task baseline.
    Modulation cells are topic hit@5 under focus/control/suppress
    instructions; capacity cells give the last-quartile plateau with the
    peak (reached at comma 20, the first block boundary, on all three
    models) in parentheses.}
  \label{tab:suite-full}
  \footnotesize
  \setlength{\tabcolsep}{3pt}
  \begin{tabular}{lccc}
    \toprule
    Family & \qwen{} 27B & \ouro{} 2.6B & \huginn{} 3.5B \\
    \midrule
    Verbal report (top-5) & 69\% (\textsc{agg}; rank-1 49\%) & 106/133 (80\%; rank-1 67\%) & 124/126 (98\%; rank-1 98\%) \\
    Introspection (8$\sigma$) & null (best rank 173 of 101) & median RR 0.50; top-1 31\% & null (top-1 0\%; 0/97) \\
    Modulation topic@5 (\textsc{agg}) & .38/.07/.07 & .77/.39/.26 & .28/.15/.19 \\
    Summoning expected@5 & 0.01 & 0.03 & 0.01 \\
    Two-hop probe swap & 47/90 (52\%); base 57/90 & 26/90 (29\%); base 43/90 & 5/90 (6\%); base 25/90 \\
    Flexible generalisation & 80/192 (42\%) & 73/192 (38\%); base 42/64 & 25/192 (13\%); base 28/64 \\
    Selectivity: language ($\Delta$@5) & +1.00 & +0.25 & +0.38 \\
    Selectivity: line count (\textsc{agg}) & 0.09/0.00 & 1.00/0.09 & 0.09/0.09 \\
    Ignition (10--90 width) & 0.10 & 0.55 & 0.76 \\
    Capacity, words held & 0.7 (peak 1.4) & 2.4 (peak 5.0) & 1.8 (peak 3.1) \\
    Dual task, of 21 (\textsc{agg}) & concept 6; maths 1 & concept 17; maths 0 & concept 7; maths 3 \\
    \bottomrule
  \end{tabular}
\end{table}

Two rows deserve notes. Summoning's zero is a \emph{no-discrimination} zero
in the strongest sense: summoned concepts score no better than matched foils
(difference $\le 0.01$ on all models) and asking versus not asking changes
nothing ($|q_1 - q_2| = 0.000$ everywhere). And the line-count selectivity
cells read ``band readability of the count when directly asked~/~when never
asked'': only Ouro shows the gated pattern.

\begin{table}[t]
  \centering
  \caption{Flexible generalisation by intervention condition --- the raw
    data behind \S\ref{sec:writes}. Each trial swaps a source concept's lens
    direction for a target concept's and scores whether the answer flips to
    the injected concept's implication; per-category columns give flips out
    of 48. Qwen's run stored only the total. No-intervention baselines:
    \ouro{} 42/64, \huginn{} 28/64 (44\%). Bottom: mean cosine between lens
    directions of same-category concepts (\ouro{}, loop-2 band) --- the
    category-separability covariate discussed in \S\ref{sec:writes}'s open
    question.}
  \label{tab:flexgen-full}
  \footnotesize
  \setlength{\tabcolsep}{4pt}
  \begin{tabular}{llccccc}
    \toprule
    & Condition & Flips/192 & \% (95\% CI) & countries & months & animals \\
    \midrule
    \qwen{}   & band L24--48 & 80 & 42 (35--49) & --- & --- & --- \\
    \midrule
    \ouro{}   & loops 2+3+4 & 73 & 38 (31--45) & 33 & 16 & 24 \\
              & loop 2 only & 40 & 21 (15--27) & 31 & 0  & 9  \\
              & loop 4 only & 48 & 25 (19--31) & 30 & 12 & 6  \\
              & loops 2+3+4, $2\times$ strength & 14 & 7 (4--11) & 10 & 0 & 4 \\
    \midrule
    \huginn{} & all three windows & 25 & 13 & 12 & 10 & 3 \\
              & last window only & 25 & 13 & 12 & 10 & 3 \\
              & second-to-last window only & 5 & 3 & 1 & 0 & 2 \\
              & all windows, $2\times$ strength & 2 & 1 & 1 & 0 & 1 \\
    \midrule
    \multicolumn{4}{l}{\ouro{} same-category direction cosine} & 0.36 & 0.67 & 0.39 \\
    \bottomrule
  \end{tabular}
\end{table}

The fourth category, numbers, is omitted from the table for width: it never
flips under any condition on either looped model (0/48 throughout; direction
cosine 0.64 on \ouro{}). Note the Huginn last-window row is not merely
equal in total to the all-windows row --- the per-category counts are
identical.

\begin{table}[t]
  \centering
  \caption{Direction ablation by condition --- the raw data behind
    \S\ref{sec:ablation}. Each cell is answers kept (unchanged and correct)
    after ablating the named direction at the intervention layers, over the
    prompts each model answered correctly at baseline. ``Early interface
    only'' is the loop-2 band on \ouro{} and the second-to-last lens window
    on \huginn{}; \qwen{} has a single interface, so the split does not
    apply.}
  \label{tab:ablation-full}
  \footnotesize
  \setlength{\tabcolsep}{4pt}
  \begin{tabular}{lccc}
    \toprule
    Ablated direction & \qwen{} ($n{=}36$) & \ouro{} ($n{=}42$) & \huginn{} ($n{=}28$) \\
    \midrule
    Argument concept, all interfaces & 26 (72\%) & 30 (71\%) & 17 (61\%) \\
    Argument concept, early interface only & --- & 42 (100\%) & 28 (100\%) \\
    Argument concept, last interface only & --- & 32 (76\%) & 17 (61\%) \\
    Answer token, all interfaces & 19 (53\%) & 8 (19\%) & 0 (0\%) \\
    Unrelated word (control) & 36 (100\%) & 42 (100\%) & 28 (100\%) \\
    \bottomrule
  \end{tabular}
\end{table}

\section{Robustness to reasoning SFT: Ouro-2.6B-Thinking}
\label{app:thinking}

We repeated the complete \ouro{} pipeline --- lens fitting on the same
1000-prompt corpus with identical hyperparameters, the structural analyses,
and the full causal suite --- on Ouro-2.6B-Thinking, the reasoning SFT of
our base checkpoint ($\sim$8.3M supervised examples of maths, code, science
and chat; the architecture is unchanged) \citep{ouro2025}. The only pipeline
change is the model identifier. One protocol note: the Thinking tokeniser
emits its reasoning prefix only when explicitly requested, and under the
suite's default chat template the model answers directly, so every
comparison below is protocol-identical to base. All numbers in this section
are re-derived by the verification suite of Appendix~\ref{app:harness}.

\paragraph{Structure is unchanged.} Deep supervision survives the SFT. The
trained exit gate still concentrates its mass on the last two loops
(0.37/0.46 on loops 3/4; base 0.34/0.49); logit-lens top-1 agreement at the
four loop-end states spans 0.62--0.90 (base 0.60--0.92); and transport still
factorises through the loop checkpoints, with mean composition cosine 0.962
over the 20 tested paths (base 0.971).

\begin{table}[t]
  \centering
  \caption{The causal suite on base \ouro{} versus its reasoning SFT.
    Conventions follow Table~\ref{tab:suite-full}; flexible-generalisation
    rows follow the per-loop conditions of Table~\ref{tab:flexgen-full}.}
  \label{tab:thinking}
  \footnotesize
  \setlength{\tabcolsep}{4pt}
  \begin{tabular}{lcc}
    \toprule
    Family & \ouro{} 2.6B (base) & Ouro-2.6B-Thinking \\
    \midrule
    Verbal report (top-5; rank-1) & 80\%; 67\% & 69\%; 55\% \\
    Introspection (8$\sigma$: top-1; median RR) & 31\%; 0.50 & 31\%; 0.50 \\
    Modulation topic@5, focus/control/suppress (\textsc{agg}) & .77/.39/.26 & .76/.45/.15 \\
    Summoning expected@5 & 0.03 & 0.03 \\
    Two-hop probe swap & 26/90 (29\%); base 43/90 & 22/90 (24\%); base 35/90 \\
    Flex-gen: no-intervention baseline & 42/64 (66\%) & 38/64 (59\%) \\
    Flex-gen: loop 2~/~3~/~4 only & 21\%~/~29\%~/~25\% & 17\%~/~28\%~/~26\% \\
    Flex-gen: loops 2+3+4 (at $2\times$ strength) & 38\% (7\%) & 35\% (5\%) \\
    Selectivity: language ($\Delta$@5) & $+$0.25 & $+$0.00 \\
    Selectivity: line count, direct/never asked (\textsc{agg}) & 1.00/0.09 & 1.00/0.18 \\
    Ignition (10--90 width, in-band) & 0.55 & 0.20 \\
    Capacity, words held (peak at comma 20) & 2.4 (5.0) & 2.1 (4.0) \\
    Capacity, pursuit inventory ($\theta \le 0.01$) & 4.9 & 5.1 \\
    Dual task, of 21 (\textsc{agg}) & concept 17; maths 0 & concept 19; maths 0 \\
    Ablation kept: arg~/~answer~/~control & 71\%~/~19\%~/~100\% & 66\%~/~26\%~/~97\% \\
    Ablation kept: loop-2 only~/~last loop only & 100\%~/~76\% & 97\%~/~76\% \\
    Pursuit-subspace ablation kept: all~/~loop-2~/~control & 0\%~/~14\%~/~88\% & 0\%~/~11\%~/~87\% \\
    \bottomrule
  \end{tabular}
\end{table}

\paragraph{All interface signatures replicate.} The two headline causal
structure results reproduce within a few points despite the SFT: writes must
still span every remaining loop (the loop-2-only clamp is half-undone by
re-derivation at 17\%, and loop-2-only ablation is again a near-complete
null at 97\% kept), and the two opposite single-loop failure modes and the
$2\times$-strength collapse all recur. Introspection is point-identical
(top-1 31\%). The pursuit-based families follow suit: the inventory
capacity plateau is 5.1 words against base 4.9, and the loop-2-only
pursuit-subspace ablation keeps 4/38 (11\%) against base 14\%, with the
all-interface condition at 0/38 and the random-atom control at 87\%. The
workspace interface is a property of the pre-trained
looped architecture, not something the reasoning SFT reorganises.

\paragraph{Behavioural results change; intervention geometry does not.} The SFT
costs a few points of raw task performance under the suite's
direct-answer protocol (flex-gen baseline 66\%$\to$59\%, verbal report
80\%$\to$69\%, probe-swap baseline 43/90$\to$35/90), and the clamped and
injected conditions move by roughly the same amount --- the ratio of
intervention effect to baseline is preserved.

\paragraph{Two notable differences.} First, the language-selectivity gate
collapses to always-on: the passage's language label is readable at ceiling
even when task-irrelevant ($\Delta$ $+$0.25 $\to$ $+$0.00), so \ouro{}'s
gated pattern in Table~\ref{tab:suite} does not replicate on the SFT model ---
consistent with reasoning SFT keeping more context features unconditionally
active in the workspace. Second, ignition sharpens: the in-band 10--90
transition width drops 0.55 $\to$ 0.20 (scrambled control 0.24 $\to$ 0.12)
--- the Thinking model commits more abruptly once the band resolves. Both
are single-condition observations; we report them as leads rather than
conclusions. Answer-direction ablation shows a smaller difference in the
opposite direction: ablating the answer
direction is slightly \emph{less} destructive on Thinking (26\% kept
vs.\ 19\%) while argument ablation is comparable.

\section{Demonstration-metric semantics}
\label{app:demo}

The metric family of \S\ref{sec:geometry} comes from the original study's
interactive demonstration, whose exact semantics are not documented; we
reverse-engineered them from its page bundle and reimplemented them over all
three models.

\paragraph{Dump protocol.} For each model we run the demonstration's own 8
chat prompts, generate a 128-token greedy completion, and record, at every
(virtual layer, position) cell, the top-8 tokens and probabilities of both
readouts: the \jlens{} ($\mathrm{softmax}(\mathrm{unembed}(h J^{\top}))$,
through the covering lens of \S\ref{sec:lens-family}) and the logit lens
($\mathrm{softmax}(\mathrm{unembed}(h))$). These grids are the released raw
artefact; every \S\ref{sec:geometry} number is a function of them.

\paragraph{Count (occupancy).} A token's count is its number of occurrences
over positions $\times$ layers $\times$ 8 slots, with whitespace-only and
non-word token strings excluded. The demonstration's default layer window
additionally skips the first 29\% of depth; \S\ref{sec:occupancy} uses full
depth instead, since the depth profile is the object of study. (The skip is
well chosen for a standard transformer: on \qwen{} it starts at L18, just
below the content onset at L20--23 that we measure independently --- but on
\ouro{} the same fraction leaves the content-free early layers of every loop
inside the window, one of the confounds that led us to drop the
demonstration's contrastive ``Diff'' statistic from this paper; its analysis
is retained in the reproduction report.)

\paragraph{Probability field.} The calibration measure of
\S\ref{sec:calibration} is the probability a readout assigns to the token
the model actually emits at the next position, averaged over positions and
prompts. Because the dumps store only each cell's top 8, a token outside
them contributes zero, making the reported values lower bounds; chance on
the $\sim$150k vocabularies is $\sim$$10^{-5}$.

\paragraph{The partial-tiling reading artefact.}
\label{app:tiling}
Under the original four-lens family, later Huginn windows sit closer to
their covering lens targets than earlier ones, and the resulting
occupancy trend rises monotonically with depth --- a curve one would
naturally read as content accumulating through the recurrences.
Completing the family to a full tiling --- twelve additional lenses, so
that a target sits at every recurrence end (v5 to v61 in steps of 4,
plus the coda) and every source lies within one recurrence of its
covering target --- flattens the trend into the plateau of
Figure~\ref{fig:cbl}. Under the top-8 readout, the only estimator dumped
under both tilings, the partial family reads 638 total slots on the
animal prompt with 32 layers at exactly zero, the full tiling 1{,}622
with none; under the inventory, 65 of 67 layers are occupied at the
primary grid point (all 67 ungated). Within each one-recurrence window
the count still rises towards the lens target --- the mean-J transport
attenuation of \S\ref{sec:bulk} made visible as a sawtooth --- but
occupancy is nonzero at every depth any lens's transport reaches.
Ouro's early-loop near-zeros, by contrast, are distance-controlled
within each loop and reflect genuine absence. With unevenly spaced lens
targets, apparent changes across depth can be caused by changing
source-to-target distance alone.

\paragraph{The animal prompt and replies.}
\label{app:transcripts}
Figure~\ref{fig:cbl} is computed over the following transcripts (the
occupancy grid covers prompt and completion positions alike). Prompt, sent
identically to all three models:
\begin{quote}\small
I'm thinking of the largest land animal. What do they eat? Answer briefly.
\end{quote}
Replies, verbatim:
\begin{quote}\small
\textbf{\qwen{}:} ``The largest land animal is the African bush elephant.
They are herbivores and primarily eat grasses, leaves, bamboo, bark, roots,
and fruits.''

\textbf{\ouro{}:} ``Elephants are the largest land animals and they eat a
variety of plants, including leaves, bark, and fruits.''

\textbf{\huginn{}:} ``The largest land animal is the African elephant,
which is a herbivore and feeds on a variety of plants, including grasses,
leaves, and bark.''
\end{quote}
The latent/spoken classification of \S\ref{sec:occupancy} follows from
these texts with singular/plural variants merged: \emph{tree} is latent for
all three models, \emph{food} for \qwen{}, \emph{mammal} and
\emph{vegetation} for \ouro{}, \emph{fruit} and \emph{root} for \huginn{}
(which, uniquely, never names them).

\section{The gradient-pursuit concept inventory}
\label{app:pursuit}

The occupancy estimator of \S\ref{sec:occupancy} is the original study's
concept inventory: a sparse non-negative reconstruction of the state from
\jlens{} atoms. This appendix gives the algorithm, the estimator's
agreement and disagreement with the demonstration's top-8 readout, and the
protocols and full results of the two pursuit-based causal families
(\S\ref{sec:ablation}, \S\ref{sec:nulls}).

\paragraph{Algorithm.} At each (virtual layer $v$, position $t$) the
dictionary is the intervention direction family of \S\ref{sec:causal},
unit-normalised: $a_\tau = J_v^{\top} u_\tau / \lVert J_v^{\top} u_\tau
\rVert$ over the vocabulary, with whitespace-only and non-word token
strings excluded exactly as in the top-8 counts
(Appendix~\ref{app:demo}). Greedy pursuit runs on the residual $r$
(initialised to $h_{v,t}$): select $\arg\max_\tau \langle r, a_\tau
\rangle$ among positive scores --- no materialised dictionary is needed,
since $\langle r, J_v^{\top} u_\tau \rangle$ is the lens-logit operation
applied to $r$, normalised by precomputed atom norms --- then refit all
selected coefficients under a non-negativity constraint (projected
gradient on the support's Gram system, batched over positions), subtract
the reconstruction, and repeat until $k_{\max} = 50$ atoms or no positive
score remains. The dumps record each step's selection, a full coefficient
snapshot at $k \in \{8, 16, 25, 50\}$, and the residual fraction, so
every occupancy count at any $(k, \theta)$ is an offline slice of one
artefact. Occupancy at $(k, \theta)$ counts the atoms among the first $k$
whose refitted coefficient is at least $\theta \lVert h \rVert$; the body
reports $k = 25$, $\theta = 0.05$, with robustness checked over the full
grid $k \in \{8, 16, 25, 50\} \times \theta \in \{0, 0.01, 0.05\}$. The
cells are the same (layer, position) grids as the top-8 dumps --- the
stored \texttt{seq\_ids} reproduce each recorded forward deterministically
--- so the two estimators align cell-for-cell on all three models.

\begin{figure}[t]
  \centering
  \includegraphics[width=\textwidth]{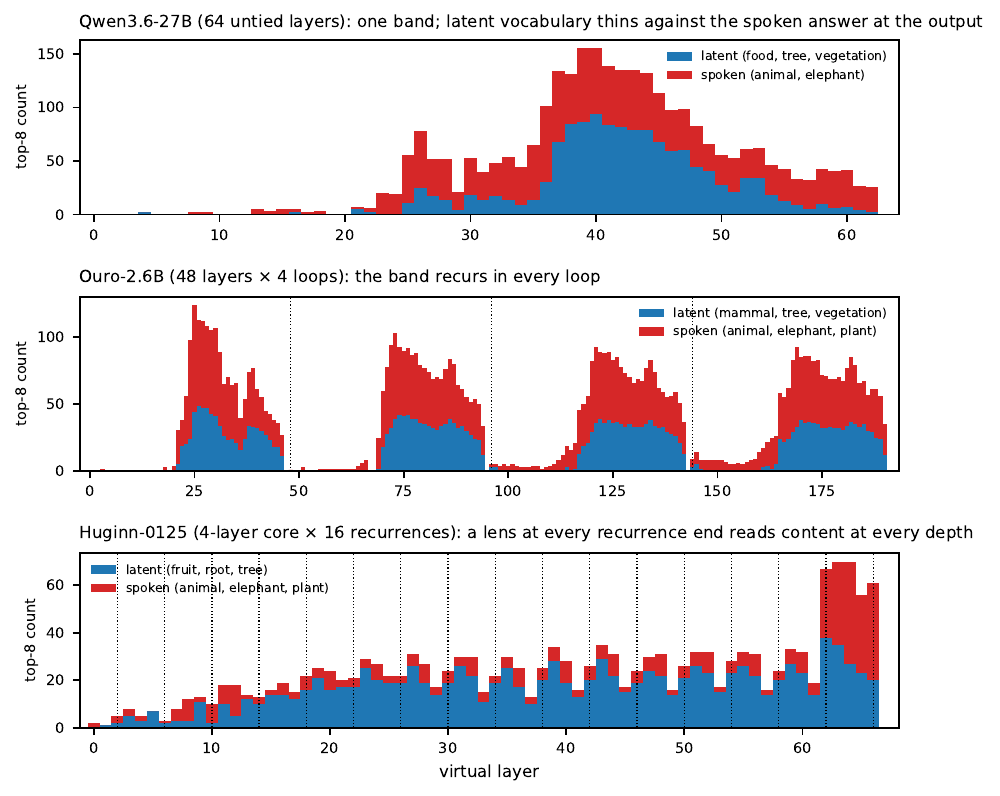}
  \caption{Count-by-layer occupancy under the demonstration's top-8
    readout --- the estimator-comparison counterpart of
    Figure~\ref{fig:cbl}, same prompt, same token classes, same layout.
    The band, loop, and window structure appears under both estimators.}
  \label{fig:cbl-top8}
\end{figure}

\paragraph{Agreement and disagreement with the top-8 readout.} The
demonstration's top-8 softmax readout --- whose semantics we
reverse-engineered from the study's page bundle
(Appendix~\ref{app:demo}) --- serves as a reference estimator: every
occupancy dump was also read through it. The two
estimators pick genuinely different tokens: mean per-cell Jaccard overlap
between the pursuit support and the top-8 slots is 0.07--0.24, and the
near-twin rate (two read tokens identical up to case or plural) falls
from 30--37\% of cells under top-8 to 0.1--0.3\% under pursuit ---
confirming the original study's remark that the inventory is less
redundant than inner-product top-$k$. Despite that, every structural
claim of \S\ref{sec:occupancy} except two holds under both estimators at
every grid point. Under top-8 the corresponding numbers are: Qwen animal
prompt, 6 latent slots below L20, 1{,}076 latent vs.\ 1{,}116 spoken in
the band, 56 vs.\ 231 at L55+; Ouro currency riddle, \emph{euro} 679
$\to$ 508 against \emph{Italy} 199 $\to$ 584 across loops; Huginn
currency riddle, 2 slots for the country family over all 67 virtual
layers (pursuit grid maximum: 9); Huginn full tiling, 1{,}622 total
slots, mid-window plateau 86--108 against 21 in the first window, coda
343. Figure~\ref{fig:cbl-top8} is the top-8 counterpart of
Figure~\ref{fig:cbl}. Two top-8 results do \emph{not} survive the
estimator change. On the animal prompt, the top-8 latent class thins
fourfold against the spoken class at L55+ (the 56 vs.\ 231 above),
whereas the inventory's latent/spoken ratio is flat-to-rising at every
grid point --- the thinning is a property of top-8's eight-slot
competition, not of the state; the layer-level commitment that is real
on Qwen lives on the sport prompt (\S\ref{sec:occupancy}). And Ouro's
in-band excess holds under top-8 on raw loop totals, but under the
inventory only as a per-layer rate (\S\ref{sec:occupancy}).

\paragraph{Variance explained.} The pursuit reconstruction at $k = 50$
explains on average 4.3\% of state variance on \qwen{}, 10.2\% on
\huginn{}, and 21.5\% on \ouro{} --- ordered exactly as the models'
decodability (\S\ref{sec:calibration}) and small in absolute terms
everywhere. The $k$-curve has no elbow: $k = 25$ captures 91\% of the
$k = 50$ value on Qwen but only 78\% on Ouro and 63\% on Huginn, so the
$k \le 25$ cap is a convention on the recurrent models, not a saturation
point. This motivates reporting occupancy with a coefficient threshold
rather than raw selection counts.

\paragraph{The coefficient gate.} Two of the \S\ref{sec:occupancy} zeros
--- Qwen's below-L20 absence on the animal prompt and its last-8-layer
sport zero --- exist only under the $\theta = 0.05\,\lVert h \rVert$
gate (exact zeros at $k \ge 16$ and $k \ge 25$ respectively; near-zero
at smaller $k$). Ungated selection fills slots at every depth: greedy
pursuit always selects \emph{something} when the atoms explain so little
of the state (4.3\% on Qwen), so absence registers as vanishing refitted
coefficients rather than as empty slots, and the gate turns ``selected
at negligible weight'' into ``not counted''. This is the role the top-8
probability floor plays implicitly for the demonstration's estimator,
which shows the same below-L20 absence (6 slots) without an explicit
threshold. On Qwen, absence and commitment are therefore statements
about coefficient mass, not about which atoms remain selectable.

\paragraph{Capacity under the inventory.} The capacity task of
\S\ref{sec:nulls} re-scored with pursuit in place of the rank probe:
identical seeded word lists and prompts, one recorded forward per trial,
pursuit at the comma positions over each model's read layers. A word
counts as held at a comma if, at any read layer, a selected atom with
coefficient $\ge \theta \lVert h \rVert$ matches it (exact token id or
case-insensitive string). Table~\ref{tab:capacity-pursuit} gives the
last-quartile plateaus at $k = 25$.

\begin{table}[t]
  \centering
  \caption{Capacity plateaus (words held, last-quartile mean) under the
    rank probe of Table~\ref{tab:suite-full} and under the pursuit
    inventory at $k = 25$ for three coefficient thresholds.}
  \label{tab:capacity-pursuit}
  \footnotesize
  \begin{tabular}{lcccc}
    \toprule
    Model & Rank probe & $\theta = 0$ & $\theta = 0.01$ & $\theta = 0.05$ \\
    \midrule
    \qwen{}        & 0.7 & 2.0 & 1.9 & 0.1 \\
    \ouro{}        & 2.4 & 4.9 & 4.9 & 4.8 \\
    \huginn{}      & 1.8 & 3.6 & 3.6 & 2.2 \\
    Ouro-2.6B-Thinking & 2.1 & 5.1 & 5.1 & 5.0 \\
    \bottomrule
  \end{tabular}
\end{table}

\paragraph{Subspace ablation.} The ten-direction ablation of
\S\ref{sec:ablation} runs in two passes. An uninstrumented forward pass
records, at every intervention (layer, position), the first ten pursuit
atoms for that cell; each cell's atoms are orthonormalised by QR into a
frozen basis $Q$. The intervention pass then applies $h \leftarrow h -
QQ^{\top}h$ at those cells --- removing the state's projection onto its
own leading inventory subspace, as computed on the clean run. The control
replaces each cell's basis with the span of ten fixed random-vocabulary
atoms (shared per layer). Trials, interfaces, and scoring are identical
to the single-direction ablation family
(Table~\ref{tab:ablation-full}). Table~\ref{tab:ablate10} gives the full
counts.

\begin{table}[t]
  \centering
  \caption{Answers kept under ten-direction pursuit-subspace ablation.
    ``Early interface only'' is the loop-2 band on \ouro{} and Thinking
    and the second-to-last lens window on \huginn{}, as in
    Table~\ref{tab:ablation-full}; \qwen{} has a single interface.}
  \label{tab:ablate10}
  \footnotesize
  \setlength{\tabcolsep}{4pt}
  \begin{tabular}{lcccc}
    \toprule
    Condition & \qwen{} ($n{=}36$) & \ouro{} ($n{=}42$) & \huginn{} ($n{=}28$) & Thinking ($n{=}38$) \\
    \midrule
    Pursuit subspace, all interfaces & 23 (64\%) & 0 (0\%) & 1 (4\%) & 0 (0\%) \\
    Pursuit subspace, early interface only & --- & 6 (14\%) & 27 (96\%) & 4 (11\%) \\
    Pursuit subspace, last interface only & --- & 0 (0\%) & 0 (0\%) & 0 (0\%) \\
    Random-atom control, all interfaces & 35 (97\%) & 37 (88\%) & 26 (93\%) & 33 (87\%) \\
    \bottomrule
  \end{tabular}
\end{table}

\paragraph{Verification.} All pursuit artefacts --- the per-cell dumps
with coefficient snapshots, the capacity dumps, and the ablation records
--- are released alongside the top-8 dumps, and a companion verification
suite (132 checks; Appendix~\ref{app:harness}) re-derives from them every
pursuit number in this paper, including each robustness verdict at all
twelve $(k, \theta)$ grid points.

\section{Worked examples}
\label{app:examples}

The body of the paper reports aggregates; this section shows single raw data
points, so the reader can see concretely what a lens readout and an
intervention record look like. Every value is pinned to its exact cell by
the verification suite of Appendix~\ref{app:harness}.

\paragraph{One prompt, three workspaces.} All three readout cells come from
the same demonstration prompt of \S\ref{sec:geometry}, whose completion must
hold a concept in mind without saying it:

\begin{quote}\itshape
Think of a specific sport, but don't say its name. Describe the atmosphere
at a big match of that sport without naming it.
\end{quote}

Table~\ref{tab:examples} shows one cell of each model's (layer $\times$
position) readout grid --- a high-probability cell chosen to be
characteristic of that model's pattern --- with the five most probable
tokens under each lens.

\begin{table}[t]
  \centering
  \caption{One readout cell per model on the hidden-sport prompt: the top-5
    tokens (with probabilities) of the \jlens{} and the logit lens at the
    same (virtual layer, position). ``While writing'' quotes the completion
    text around the position being read. \textvisiblespace{} marks a
    tokeniser leading space. The three cells show the three regimes the
    paper describes: on \qwen{}, the unspoken concept is sharp in the
    \jlens{} and invisible to the logit lens; on \ouro{}, a mid-loop state
    read through that loop's own lens holds the candidate sports the model
    is still deliberating; on \huginn{}, both lenses read the \emph{same}
    word --- the \jlens{} sharpens (0.88) what the near-token-space state
    already shows weakly (0.06).}
  \label{tab:examples}
  \footnotesize
  \setlength{\tabcolsep}{3pt}
  \begin{tabular}{p{3.6cm}lp{8.2cm}}
    \toprule
    Cell & Lens & Top-5 readout \\
    \midrule
    \qwen{} L42, & \jlens{} & \texttt{\textvisiblespace football} .54,
      \texttt{\textvisiblespace soccer} .23,
      \texttt{\textvisiblespace Football} .09, \texttt{Football} .04,
      \texttt{football} .03 \\
    ``thud of a boot connecting with a [sphere]'' & logit &
      \texttt{\textvisiblespace(} .00, \texttt{...} .00,
      \texttt{\textvisiblespace bag} .00, \texttt{sc} .00,
      \texttt{\textvisiblespace**} .00 \\
    \midrule
    \ouro{} loop 3, physical L26, & \jlens{} &
      \texttt{\textvisiblespace football} .50,
      \texttt{\textvisiblespace soccer} .21,
      \texttt{\textvisiblespace sports} .09,
      \texttt{\textvisiblespace rugby} .07,
      \texttt{\textvisiblespace stadium} .03 \\
    ``describe the atmosphere at a [big]'' & logit &
      \texttt{\textvisiblespace Football} .03,
      \texttt{\textvisiblespace football} .02,
      \texttt{\textvisiblespace championship} .01,
      \texttt{\textvisiblespace stadium} .01,
      \texttt{\textvisiblespace sports} .01 \\
    \midrule
    \huginn{} recurrence 15, & \jlens{} &
      \texttt{\textvisiblespace excitement} .88,
      \texttt{\textvisiblespace energy} .05,
      \texttt{\textvisiblespace emotion} .03,
      \texttt{\textvisiblespace enthusiasm} .01,
      \texttt{\textvisiblespace emotions} .01 \\
    ``stadium was alive, pulsating with [the]'' & logit &
      \texttt{\textvisiblespace excitement} .06,
      \texttt{\textvisiblespace intensity} .04,
      \texttt{\textvisiblespace energy} .02,
      \texttt{\textvisiblespace anticipation} .02,
      \texttt{\textvisiblespace electricity} .01 \\
    \bottomrule
  \end{tabular}
\end{table}

Each cell illustrates a characteristic readout pattern for one model. \qwen{} is writing
``\ldots{}the heavy thud of a boot connecting with a \emph{sphere}'' ---
deliberately avoiding the word \emph{ball} --- and at mid-band L42 the
\jlens{} reads the sport it is not naming at 0.54, while the logit lens at
the same state reads punctuation fragments at $\sim$$10^{-3}$: mid-stack
content exists but is not token-decodable without transport
(\S\ref{sec:calibration}). \ouro{}'s cell sits mid-loop --- between the
checkpoints deep supervision trains --- while its completion is still the
generic ``the user wants me to describe the atmosphere at a\ldots''; the
loop-3 lens reads the candidate sports the model is weighing (it later
settles on rugby), where the same state's direct decode is an order of
magnitude weaker. \huginn{}'s cell shows why we call its states
near-token-space (\S\ref{sec:implicit}): the logit lens already reads the
right word everywhere, weakly, and the \jlens{} concentrates it.

\paragraph{One intervention record.} The introspection experiment
(\S\ref{sec:verbalisation}) injects a concept the model is \emph{not}
computing along its lens direction, asks the model to name what it is
thinking of, and records the injected concept's rank in the answer
distribution at each injection strength. The per-trial records make single
concepts inspectable; the concept \emph{lightning} appears in all three
models' concept lists:

\begin{center}
  \footnotesize
  \setlength{\tabcolsep}{6pt}
  \begin{tabular}{lcccccc}
    \toprule
    Injection strength & $0\sigma$ & $0.5\sigma$ & $1\sigma$ & $2\sigma$ & $4\sigma$ & $8\sigma$ \\
    \midrule
    \ouro{}   & 7{,}524  & 1{,}809 & 622    & 114    & 31     & \textbf{8} \\
    \qwen{}   & 10{,}533 & 1{,}084 & 533    & 292    & 218    & 189 \\
    \huginn{} & 15{,}969 & 3{,}755 & 3{,}935 & 3{,}798 & 3{,}422 & 3{,}151 \\
    \bottomrule
  \end{tabular}
\end{center}

The verbalisation split of \S\ref{sec:verbalisation} is visible in this one
row. As injection strength increases, the concept's rank improves
monotonically in \ouro{} (explicit per-iteration supervision) until it enters
reportable range. The rank also improves substantially in \qwen{} but remains
two orders of magnitude below that range. In \huginn{}, the rank responds to
the first increment and then plateaus: the injection measurably moves the
state, but nothing carries it to the output as a word.

\section{Adapter validation and silent-failure traps}
\label{app:traps}

Fitting a Jacobian lens through a recurrence exposes failure modes that do not
exist on feedforward stacks. Each of the following corrupts results
\emph{without raising an error}:

\begin{enumerate}
  \item \textbf{No-grad recurrence phases.} Huginn's iterated forward splits
    recurrences into no-grad and with-grad phases; passing a scalar step count
    places \emph{all} of them in the former, making every Jacobian through the
    recurrence exactly zero. The fix is to request $(0, r)$ steps; our smoke
    test asserts a non-zero Jacobian at every recurrence.
  \item \textbf{Stochastic state initialisation.} Huginn draws its initial
    recurrent state afresh per forward. The adapter seeds the generator per
    call (and disables test-time noise); determinism is verified bit-exact,
    since a lens averaged over unmatched forwards is an average over different
    computations.
  \item \textbf{Gradient checkpointing.} Backward-pass recompute re-fires
    block hooks and desynchronises the virtual-layer counters; checkpointing
    must stay off.
\end{enumerate}

The adapter's smoke tests additionally verify that each physical layer yields
distinct per-firing activations, that the virtual-final state reproduces the
model's own logits exactly, and that forwards are bit-exact deterministic
under the seeded initialisation.

\section{transformers-5 compatibility}
\label{app:compat}

Both looped models ship remote code written against transformers~4.x; our
experiments run under 5.14. \ouro{} needs three load-time patches: (a)
\texttt{config.pad\_token\_id} is no longer auto-created and must be set to
\texttt{None}; (b) the rotary-embedding registry lost its \texttt{'default'}
(classic RoPE) entry, so a shim re-registers the default parameter function
on the dynamically loaded rotary class; (c) \texttt{create\_causal\_mask}
renamed keyword arguments. \huginn{} needs exactly one:
it declares \texttt{\_tied\_weights\_keys} in the 4.x list form, which
crashes 5.x \texttt{post\_init}; the adapter rewrites it to the dict form
\texttt{\{"lm\_head.weight": "transformer.wte.weight"\}} before
\texttt{from\_pretrained}. Huginn's custom recurrence cache is untested under
5.x; no experiment in this paper uses it (all forwards are cache-free and
nothing calls \texttt{generate()}).